\pdfoutput=1

\documentclass[11pt,a4paper]{article}
\usepackage{amsmath,amsfonts}
\usepackage{algorithmic}
\usepackage{algorithm}
\usepackage{array}
\usepackage[caption=false,font=normalsize]{subfig}
\usepackage{textcomp}
\usepackage{stfloats}
\usepackage{url}
\usepackage{verbatim}
\usepackage{tabularx}
\usepackage{graphicx}
\usepackage{cite}
\usepackage{appendix}

\usepackage{lipsum}

\begin{document}

\title{Conformal Recursive Feature Elimination}

\author{Marcos López-De-Castro, Alberto García-Galindo, Rubén Armañanzas
\thanks{\textit{Marcos López-De-Castro, Alberto García-Galindo and Rubén Armañanzas are with DATAI - Institute of Data Science and Artificial Intelligence, Universidad de Navarra, Pamplona, Spain and Tecnun School of Engineering, Universidad de Navarra, Donostia-San Sebastian, Spain. Email:\{mlopezdecas, agarciagali, rarmananzas\}@unav.es}}}


\maketitle
\footnotetext[1]{Correspondig: mlopezdecas@unav.es }
\footnotetext[2]{Under review}

\begin{abstract}
Unlike traditional statistical methods, Conformal Prediction (CP) allows for the determination of valid and accurate confidence levels associated with individual predictions based only on exchangeability of the data. We here introduce a new feature selection method that takes advantage of the CP framework. Our proposal, named Conformal Recursive Feature Elimination (CRFE), identifies and recursively removes features that increase the non-conformity of a dataset. We also present an automatic stopping criterion for CRFE, as well as a new index to measure consistency between subsets of features. CRFE selections are compared to the classical Recursive Feature Elimination (RFE) method on several multiclass datasets by using multiple partitions of the data. The results show that CRFE clearly outperforms RFE in half of the datasets, while achieving similar performance in the rest. The automatic stopping criterion provides subsets of effective and non-redundant features without computing any classification performance.
\end{abstract}

\vspace{0.5cm}
\vspace{0.5cm}

\section{Introduction}

The curse of dimensionality is a well-known issue in the field of statistical learning theory. In recent years, the amount of high-dimensional and multi-modal data has become a challenge, from healthcare \cite{Berisha_2021, andreu-perez_2015} to  physics \cite{Wright_2014}, as a consequence of technological advances and the big data advent. Feature selection methods are techniques developed for dimensionality reduction that select an optimal subset of features without altering their original meaning \cite{jain_1997}. The use of these methods helps prediction algorithms to perform faster and to increase their efficiency. These techniques are categorized as filters, wrappers, and embedded methods, although mixed approaches have also been proposed \cite{Guyon_2006, Chandrashekar_2014}. Search strategies have been developed to explore the feature space efficiently \cite{data-driven_2020}. Two outstanding techniques are the sequential forward and the sequential backward selection. They are based on adding and removing features until the required number of features is satisfied, respectively.
The Recursive Feature Elimination method (RFE) is a popular feature selection method proposed by Guyon \textit{et al.} based on a backward elimination policy \cite{Guyon_et_al_2002, zhou_et_al_2007}. RFE was originally developed for support vector machines and performs by recursively removing the features, or sets of features, that least decreases the margin of separation between classes.
\\

In this work, we introduce a novel recursive feature selection method based on the RFE algorithm for the Conformal Prediction framework.
When predictions are performed in high-risk scenarios, confidence levels concerning individual predictions are sought \cite{Nouretdinov_2011, Panch_2019}. Conformal Prediction provides a powerful and innovative framework able to offer non-asymptotic theoretical guarantees for uncertainty quantification in individual predictions \cite{ALRW, CP4RML, Anastasios_2023, chen_2023}. This new framework is independent of the learning rule implemented and only requires the Independent and Identically Distributed (i.i.d.) nature of data. Consequently, Conformal Prediction is the appropriate tool for high-risk machine learning applications.
Few methods to accomplish feature selection within this framework have been proposed. One of these, developed by Bellotti \textit{et al.} \cite{Bellotti_et_al_2006}, is \textit{`Strangeness minimization Feature Selection'} (SMFS). The idea behind the SMFS method is to create a ranking of features and choose those that contribute less to global strangeness.
Yang \textit{et al.} proposed the \textit{`Average Confidence Maximization'}  (ACM) method \cite{Yang_2011}. It is based on computing the confidence level by taking into account only individual features. The feature that maximizes the confidence is selected and then the process is repeated by adding features until achieving the required confidence level. Cherubin \textit{et al.} \cite{Cherubin_2018} used the i.i.d assumption to select relevant features in an anomaly detection problem and Zhou \textit{et al.}  presented a conformal feature-selection wrapper for instance transfer \cite{Zhou_2018}. To the best of our knowledge, no other relevant feature selection method in Conformal Prediction have been developed. \\

We propose Conformal Recursive Feature Elimination (CRFE), a new feature selection method that combines strangeness with the backward recursive elimination policy. As features are removed, the strangeness between samples is updated, similarly to how feature weights are updated in the original RFE. The features removed are associated with higher strangeness. We compared CRFE and RFE results on four different multiclass datasets. CRFE outperformed RFE results in two of the four datasets and achieved the same performance and consistency as RFE in the remaining.
We also present an automatic stopping criterion and a new consistency index. The automatic stopping criterion stops the recursive elimination of features, avoiding full dataset computation.\\

The rest of the manuscript is organized as follows. Section \ref{chap_2} briefly introduces the conformal prediction framework. Section \ref{chap3} develops the tools needed to support the multi-class scenario, presents the recursive algorithm and the stopping criteria. Experimental comparison is described in Section \ref{chap4} and results are detailed and discussed in Section \ref{chap5}. Section \ref{chap6} concludes the paper.

\section{Conformal prediction} \label{chap_2}

Conformal prediction framework rests in two assumptions: randomness and exchangeability \cite{Toccaceli_2022}. A finite sequence $(z_1, ..., z_n)$ of random variables is said to be i.i.d, \textit{i.e.}, the randomness assumption, if they are generated independently from the same unknown probability distribution $Q$. A finite sequence $(z_1, ..., z_n)$ of random variables is said to be exchangeable if for any permutation $\pi$ of the set of index $\{1,...,n\}$ the joint probability distribution is invariant $Q(z_1, ..., z_n)= Q'(z_{\pi_{1}}, ..., z_{\pi_{n}})$. Randomness assumption implies exchangeability. The construction of a \textit{non-conformity measure} is the next step when conformal prediction is performed. A non-conformity measure is defined as a measurable function $A$,
\begin{align*}
  A:\mathbb{R} \times \mathbb{R}^{n} &\to \mathbb{R}\\
  (z, \{z_1,...,z_n\}) &\mapsto \alpha,
\end{align*}
that quantifies the degree of strangeness of a sample $z$ with respect to a bag of samples $\{z_1, . . . , z_n\}$. The larger \(\alpha\) is, the stranger \(z\) is with respect to \(\{z_1, . . . , z_n\}\). The function \(A\) must be invariant with respect to any ordering of  $\{z_1, . . . , z_n\}$. When a classification task is conducted, a natural choice for the non-conformity function is
\begin{equation}
    \label{class_NCF}
    A(z,\{z_1,...,z_n\}) = f(y,h(x)),
\end{equation}
where $h(\cdot)$ is a predictive rule learned from $\{z_1,...,z_n\}$ and $f(\cdot,\cdot)$ a function which quantifies the difference between the prediction $h(x)$ and a label $y$. Monotonic transformations of $f(\cdot,\cdot)$ have no impact on the predictions. Nevertheless, the proper selection of the estimator $h(\cdot)$ has a significant effect in conformal prediction efficiency \cite{Shafer_2008}. Let’s assume a set of samples of the form 
\begin{equation}\label{set_form}
\{z_i\}^n_{i=1} = \{(\mathbf{x}_i, y_i)\}^n_{i=1} \in \mathcal{X} \times \mathcal{Y} = \mathcal{Z}. 
\end{equation}
A \textit{confidence predictor} $\Gamma^{\epsilon}$ is defined as the prediction interval that will contain the true label of a new sample with a confidence level $\epsilon \in [0, 1]$,
\begin{equation} 
    P(y_{n+1} \in \Gamma^{\epsilon}(\mathbf{x}_{n+1}) ) \geq 1 - \epsilon.
    \label{validity}
\end{equation}
The prediction interval is then defined as
\begin{equation}
\Gamma^{\epsilon}(\mathbf{x}_{n+1})=\{\; y\; | \;p^{(\mathbf{x}_{n+1},y)} > \epsilon\;\} \;\;\;\;\; \forall y\in \mathcal{Y},
\label{pred_set}
\end{equation}
where the p-values \(p^{(\mathbf{x}_{n+1},y)}\) are usually derived following a \textit{transductive} or an \textit{inductive} approach. When transductive inference is followed, all computations have to start from scratch. In particular, for every test sample the transductive inference needs to learn  $|\mathcal{Y}|$ prediction rules $h(\cdot)$ to build the confidence sets (\ref{pred_set}). Inductive conformal prediction was developed to deal with this computational inefficiency \cite{Papadopoulos_2008}. Inductive approach splits the set of samples into a training $\{z_1, ..., z_l\}$  and calibration $\{z_{l+1}, ..., z_n\}$ sets. A general prediction rule $h(\cdot)$ is learned from the training set and the p-values in (\ref{pred_set}) are computed exclusively from the calibration set as
\begin{equation}
    p^{(\mathbf{x}_{n+1},y)} = \frac{|\{ \;i = l+1,...,n+1 \; | \;\alpha_i \geq \alpha_{n+1}^y \;\}| + 1}{n-l+1}.
    \label{p_values}
\end{equation}
The general rule inducted from the training set incorporates the information about the dataset, eliminating the need to start from scratch when the confidence sets for the new test samples are derived. Hence, inductive inference is almost as computationally efficient as the implemented prediction rule is. The inductive approach sacrifices prediction efficiency for computational efficiency due to the train-calibration split.

\section{Methodology} \label{chap3}
In this section CRFE is presented. First, we adapt the mathematical tools developed by Belloti \textit{et al}. \cite{Bellotti_et_al_2006} to fit multi-class classification, where the potential of conformal prediction becomes particularly relevant. This adaptation ties non-conformities with features. After this, CRFE is presented. Finally, we propose a stopping criterion based on the behavior of the selected features non-conformities.

\subsection{The multiclass adaptation}

Our feature selection method is model-dependent. It is suitable for those classifiers whose decision function can be expressed in terms of a linear discriminant function. 
Let's start by assuming that the set of samples \(\mathcal{Z}\) in (\ref{set_form}) is composed by \(l\) features and \(m\) different classes, \textit{i.e.},
\begin{align*}
    \mathbf{x}_i \in \mathcal{X} = (X_1,...,X_l),\\ y_i \in  \mathcal{Y}= \{y^{ 1},..., y^{ m} \},
\end{align*}
where \(X_j \in \mathbb{R}\).
We follow the One vs All (OVA) approach \cite{zhou_et_al_2007} for extending \cite{Bellotti_et_al_2006} to a multiclass scenario. Let define the function \(\theta\) as:
\begin{equation}
    \theta(y_i,y^{ k }) = \left\{ \begin{array}{lcc}
             1 &   if  & y_i = y^{ k }  \\
             \\ -1 &   if  & y_i \neq y^{ k }
             \end{array}
   \right. 
\end{equation}
where \(k \in \{1, ..., m\}\), such that we can derive \(m\) new datasets to train \(m\) linear models in the OVA approach:
\begin{align}
\mathcal{Z}^k =  \{z_i^k\}_{i=1}^n
= \{(\mathbf{x}_i,\theta(y_i,y^{ k }))\}_{i=1}^n.
\end{align}
When training the classifier on each dataset, \(h_k = h(\mathcal{Z}^k)\), \(m\) weight vectors \(\mathbf{w}^k\) defining the separation hyperplanes are produced.
In the original binary classification task, a reasonable non-conformity function for this kind of classifiers was proposed by Vovk \textit{et al.} \cite{ALRW}:
\begin{equation}
\label{ref_2}
    \alpha_i^{binary} = \tilde{A}(z_i^k,\mathcal{Z}^k,h_k) = - \theta(y_i,y^{ k }) \sum_{j=1}^l (w_j^k x_{ij} + b).
\end{equation}
Similarly to the weighted average method \cite{ALRW}, we extend the binary non-conformity measure to the multiclass problem as
\begin{align}
\label{ref_1}
    & \alpha_i = A(z_i,\mathcal{D}) \nonumber = \\ & \lambda \tilde{A}(z_i^k,\mathcal{Z}^k,h_k) + \lambda^{\prime} \sum^m_{\substack{r=1\\r\neq k}} \tilde{A}(z_i^r,\mathcal{Z}^r,h_r)  \\ & \nonumber \text{with} \; k \in \{1, ..., m\}, \; \lambda\in [0,1], \;  \text{and} \; \lambda^{\prime} = \frac{1-\lambda}{m-1},
\end{align}
where $\alpha_i$ is the non-conformity measure of the sample $z_i$.\\
Let define the subset of features \(S\) as \(\{\mathcal{X}^\prime : \mathcal{X}^\prime \subseteq \mathcal{X} = (X_1,...,X_l)\) where \(|S| = t\)\}. If we select a subset of $t$ features $S$ from the original set of features \(\mathcal{X}\), the expression defined in (\ref{ref_2}) turns into
\begin{align}
    &\alpha_i^{ S; \; binary} =  \tilde{A}(z_i^k,\mathcal{Z}^k,h_k) =  \nonumber \\& - \theta(y_i,y^{ k }) \sum_{j \in S} (w_j^k x_{ij} + b),
\end{align}
which is linear separable by features \cite{Bellotti_et_al_2006} (\textit{i.e.}, we can isolate by features the terms that are feature dependent).\\

The OVA approach of the problem defined in (\ref{ref_1}) is still linearly separable. Considering the non-conformity measure (\ref{ref_2}), the expression in (\ref{ref_1}) is expanded as
\begin{align}
\label{multi}
&A(z,\mathcal{Z}, S) = \nonumber \\ & \nonumber -\lambda \theta(y,y^{ k }) (\sum_{j \in S} w_j^k x_{j} + b^k) - \\ \nonumber &\lambda^{\prime}\sum^m_{\substack{r=1\\r\neq k}}\theta(y,y^{ r }) (\sum_{j \in S} w_j^r x_{j} + b^r)  = 
\\ \nonumber
-\sum_{j \in S}&  \Biggr[ \lambda \theta(y,y^{ k })w_j^k  +   \lambda^{\prime}\sum^m_{\substack{r=1\\r\neq k}}\theta(y,y^{ r })  w_j^r\Biggr] x_{j} -  \gamma, 
\end{align}
where the constant \(\gamma = \lambda \theta(y,y^{ k }) b^k + \lambda^{\prime}\sum^m_{\substack{r=1\\r\neq k}}\theta(y,y^{ r })b^r \).
\\
Our non-conformity function can then be separated by features,
\begin{equation*}
    A(z, \mathcal{D}, S) = \sum_{j \in S} \phi(z,\mathcal{D},j),
\end{equation*}
satisfying the condition of linearity.
As consequence, we can define the $\beta$-measures as 
\begin{equation*}
    \beta_j = \sum_{i=1}^n \phi(z,\mathcal{D}, j),
\end{equation*}
where
\begin{align}
\label{bj}
     &\beta_j = -  \Biggr[\lambda \sum_{i=1}^n w_j^k  \theta(y_i,y^{\prime \; k})x_{ij} \;+ \nonumber \\ & \lambda^{\prime} \sum^m_{\substack{r=1\\r\neq k}} \sum_{i=1}^n w^r_j \theta(y_i,y^{\prime \; r})x_{ij}\Biggr] - \sum_{i=1}^n \gamma_i.
\end{align}

\subsection{Conformal recursive feature elimination}

Following the idea that \textit{`a good feature ranking criterion is not always a good feature subset criterion'} \cite{Guyon_et_al_2002}, the feature selection method we present is based on a recursive backward elimination policy. We propose the recursively removal of the most non-conformal feature and to re-train the classifier in each iteration until a stopping criterion is met. The recursive algorithm can be summarized as follows:
\begin{enumerate}
    \item Train the classifier.
    \item Compute $\beta_j$ for each feature.
    \item Remove the feature $X_j$ with the higher $\beta_j$.
    \item Retrain with the new subset or stop if the stopping criterion is met.
\end{enumerate}  
Next, we show that the variation in the non-conformity function is proportional to $\beta_j$ when the feature \(X_j\) is eliminated. We use the binary framework (\ref{ref_2}) without loss of generality.
Let define $\delta A^{j}$ as the variation of the linear non-conformity function  resulting from the removal of feature $j$ in a particular calibration set:

\begin{gather*}
    \delta A_i^{k,j} = \delta A(z_i^k,\mathcal{Z}^k,h_k)^{j} = \alpha_i - \alpha_i^j =\\ - y_i\sum_{p=1}^l (w_p^k
    x_{ip} + b) +  y_i\sum^l_{\substack{s=1\\s\neq j}} (w_s^k x_{is} + b) = \\ - y_i w_j x_{ij},
\end{gather*}
where $\alpha_i^j$ is the non-conformity measure without taking into account the feature $j$.
In the whole calibration set, the variation will be :
\begin{equation}
    \delta A^{j} = \sum^n_i \delta A_i^{k,j} =   \sum^n_i - y_i w_j x_{ij}. 
\end{equation}
However, we show that
\begin{align*}
\sum^n_i - y_i w_j x_{ij} = & -  w_j \sum^n_i y_i x_{ij},
\end{align*}
so
\begin{equation}
    \label{key}
     \delta A^{j} = \beta_j.
\end{equation}
The result in (\ref{key}) strengths the interpretation of \(\beta_j\) as the non-conformity associated to a feature and justifies the fact that eliminating the feature \(X_j\) with the highest associated non-conformity is what contributes the most to reducing the global non-conformity of the calibration set. Our goal is to update the predictive rule \(h(\cdot)\) (\ref{class_NCF}) in order to produce more representative non-conformity measures. Moreover, when comparing the proposed algorithm with the original RFE method, only one additional step is included: computing expression (\ref{bj}). This additional step adds a linear computational complexity of \(\mathcal{O}(lmn)\), or \(\mathcal{O}(ln)\) if we have only two classes, over the RFE algorithm complexity.

\subsection{Stopping criterion} \label{Sub_stopp}
The optimal number of features to be selected is not a straightforward problem when feature selection is performed.
We consider two stopping criteria in our study. The first criterion (i) involves fixing a certain number of features and leave the proposed method to iterate up to an specified number of features. In cases where the optimal number of features is known this criterion is useful. Moreover, when the dataset is not too large, this criterion allows the exploration of all possible subset sizes and analyze performance metrics in order to discover an optimal number of features. However, for large datasets, this approach can become impractical. The second stopping criterion  (ii) is a novel approach 
that takes advantage of the relative $\beta$-measures variation during the recursive elimination method. The underlying idea behind this criterion is as follows: when a feature $X_j$ is removed from the set of remaining features, the resulting non-conformity associated to the new set should have decreased. We found that the mean of the $\beta$-measures decreases with constant rate until an exponential decay is observed. We propose to stop the selection process before reaching the exponential behavior. Because the transition between the constant and the exponential rate involves a deceleration, we use the second derivative of the mean, which remains close to zero until the exponential behavior starts. The exponential regime starts when the value of the second derivative exceeds at least three times the standard deviation of the set of values corresponding to the previous second derivatives. The standard deviation can be computed based on the \(k\)-latest values to enhance computational performance. The full method is presented in Algorithm \ref{alg1}.

\begin{algorithm}[H]
\caption{CRFE and $\beta$-based stopping criteria. }\label{alg:alg1}
\begin{algorithmic}
\STATE
\STATE \textbf{Input: } $\mathcal{D}_{train} = (\mathbf{X}_{train}, \mathbf{Y}_{train})$; training set
\STATE \textbf{Input: } $\mathcal{D}_{cal} = (\mathbf{X}_{cal}, \mathbf{Y}_{cal})$; calibration set
\STATE \textbf{Input: } $\sigma;$ confidence, $\ge$ than 3 is recommended
\STATE \textbf{Input: } $\psi;$ lenght of set used to compute std
\STATE \textbf{Initialize array to zero: } \textit{\textbf{beta\_means}}
\STATE \textbf{Initialize array to zero: } \textit{\textbf{beta\_num\_der}}
\STATE 
\STATE  \textbf{Do} 
\STATE \hspace{0.5cm} \(\mathbf{w},b \gets \) {\textsc{train }}\((\mathcal{D}_{train})\) 
\STATE \hspace{0.5cm} \(\boldsymbol{\beta} \gets \) {\textsc{compute }}\((\mathcal{D}_{cal}, \mathbf{w}, b)\) 
\STATE \hspace{0.5cm} \(j \gets \) {\textsc{index(max( }}\(\boldsymbol{\beta}\) ))
\STATE 
\STATE \hspace{0.5cm} $\backslash$ $\backslash$ Stopping criteria evaluation  
\STATE
\STATE \hspace{0.5cm} \textbf{If} len(\textit{\textbf{beta\_means}})  \(>\psi\):
\STATE \hspace{1.0cm}  {\textsc{delete the older element(\textit{\textbf{beta\_means}})}} 
\STATE
\STATE \hspace{0.5cm} \(\overline{\beta} \gets \)
{\textsc{Compute mean ( \(\boldsymbol{\beta}\) )}}
\STATE \hspace{0.5cm} \textit{\textbf{beta\_means}} \(\gets\){\textsc{append to array (\(\overline{\beta}\)) }} 
\STATE \hspace{0.5cm} \(\boldsymbol{x} \gets \) {\textsc{ array of integers (0,len(\textit{\textbf{beta\_means}}))}}
\STATE \hspace{0.5cm} \(\boldsymbol{\beta}^{\prime\prime} \gets \) {\textsc{derive(derive( \textit{\textbf{beta\_means}}, \(\boldsymbol{x}\) ), \(\boldsymbol{x}\) )}}  
\STATE \hspace{0.5cm} \(\beta^{\prime\prime}_{\sigma} \gets \)  {\textsc{compute std(\textit{\textbf{beta\_num\_der}})}}
\STATE \hspace{0.5cm} \textit{\textbf{beta\_num\_der}} \(\gets\)   {\textsc{append to array(\(\boldsymbol{\beta}^{\prime\prime}\))}}
\STATE
\STATE \hspace{0.5cm} \textbf{If}{\textsc{ abs(
\(\boldsymbol{\beta}^{\prime\prime}\))  \(<\)  abs(\(\sigma\beta^{\prime\prime}_{\sigma} \) ):
  }}
\STATE \hspace{1.0cm}  {\textsc{stopping criteria is met =}} \textit{true}
\STATE
\STATE \hspace{0.5cm}  {\textsc{delete feature }} \textit{j} {\textsc{from }} \(\mathbf{X}_{train}\)
\STATE \hspace{0.5cm}  {\textsc{delete feature }} \textit{j} {\textsc{from }} \(\mathbf{X}_{val}\)
\STATE 
\STATE  \textbf{Until}  {\textsc{stopping criteria is met =}} \textit{true}
\STATE
\STATE \textbf{Output: }\textit{Set of selected features}
\STATE 
\end{algorithmic}
\label{alg1}
\end{algorithm}

\section{Experimental settings}\label{chap4}

\subsection{Datasets}
Four publicly available databases were used as test-bed: one synthetic and four real-world datasets. Their characteristics are summarized in Table \ref{datasets}.
\begin{table*}[h]
\label{datasets}
\centering
\resizebox{1.02\textwidth}{!}{
\begin{tabular}{lccccccc}
\hline
\hline
\textbf{Dataset} & \textbf{Samples} &  \textbf{Classes}  &  \textbf{Features} & \textbf{Distribution of classes}  &  \textbf{Reference}  \\
\hline
\textbf{Synthetic} & 350  & 4 & 35 & (25.0, 25.0, 25.0, 25.0)  &  \cite{scikit-learn}  \\ 
\textbf{Coronary artery} &  899 & 4 & 32 & (44.9, 21.2, 14.5, 19.3)    &  \cite{UCI}  \\ 
\textbf{Dermatology} & 366 & 6 & 34 & (30.6, 16.7, 19.7, 13.4, 14.2, 5,5)   & \cite{UCI}
 \\
\textbf{Myocardial infarction} & 1700 & 8 & 104 & (84.1, 6.5, 1.1, 3.2, 1.3, 0.7, 1.6, 1.6)  &  \cite{Golovenkin_2020} \\
 \hline
 \hline
\end{tabular}
}
\caption{Datasets description after processing the data.}
\end{table*}
Data pre-proccessing involved (i) one-hot class encoding and cleaning of  features, (ii) checking for missing values; if more than 25\% of the values were missing, the feature was removed, (iii) imputation of missing values, and, (iv) standardization of the data to avoid scale biases.
\begin{itemize}
    \item \textit{\textbf{Synthetic dataset.}}  We considered a synthetic dataset to control informative versus noisy features. The designed dataset comprises 10 informative and 25 noise features. We included 4 classes distributed equally among samples, except by a 5\% which were assigned randomly.
    
    \item \textit{\textbf{Coronary artery disease dataset.}} This dataset comes from a real-world clinical problem focused on coronary artery disease diagnosis \cite{UCI, Detrano_1969}. Class 0 represents the absence of disease, whereas classes 1, 2, 3, and 4 represent the degree of artery disease. Due to the limited number of samples in class 4, we merged samples from classes 3 and 4. 
    
    \item \textit{\textbf{Dermatology dataset.}} The third dataset aims to determine the type of \textit{erythemato-squamous} disease that a group of patients are suffering from \cite{UCI, Altay_1998}.
    Included classes are \textit{Psoriasis}, \textit{Seborrheic Dermatitis}, \textit{Lichen Planus}, \textit{Pityriasis Rosea}, \textit{Chronic Dermatitis}, and \textit{Pityriasis Rubra Pilaris}.
    
    \item \textit{\textbf{Myocardial infarction dataset.}} The last dataset deals with myocardical infractions diseases \cite{Golovenkin_2020}.
    The proposed classification task is to predict if a patient will survive and, if not, the cause of death. The classes considered were: \textit{alive}, \textit{cardiogenic shock}, \textit{pulmonary edema}, \textit{myocardial rupture}, \textit{progress of congestive heart failure}, \textit{thromboembolism}, \textit{asystole}, and \textit{ventricular fibrillation}.
\end{itemize}
We encourage the reader to review the \textit{Supplementary Material} for additional details.

\subsection{Performance evaluation}
Conformal prediction performance, \textit{i.e.}, set prediction results, was evaluated using the following metrics \cite{Vovk_2016}:
\begin{itemize}
    \item \textbf{Coverage}: Empirical coverage measures the percentage of test samples for which the true class falls in the prediction set. It must be close to the theoretical coverage \(1-\epsilon\). For \(k\) test samples,
    \begin{equation*}
        Cov = \frac{1}{k} \sum_{i=1}^{k} \mathbf{1}_{\{y_i \in \Gamma_i^{\epsilon}(x_i)\}}.
    \end{equation*}
    
    \item \textbf{Inefficiency (N-Score)}: This performance score measures the average size of the prediction sets:
    \begin{equation*}
        Ineff = \frac{1}{k} \sum_{i=1}^{k} |\Gamma_i^{\epsilon}(x_i)|.
    \end{equation*} 
    \item \textbf{Certainty}: This score is defined as the percentage of test samples correctly classified and with a prediction set of size 1:
    \begin{equation*}
        Cert = \frac{1}{k} \sum_{i=1}^{k} \mathbf{1}_{\{\Gamma_i^{\epsilon}(x_i) = {y_i}\}}.
    \end{equation*}
    \item \textbf{Uncertainty}: Proportion of test samples for which the prediction set includes all classes:
    \begin{equation*}
        Uncert = \frac{1}{k} \sum_{i=1}^{k} \mathbf{1}_{\{\Gamma_i^{\epsilon}(x_i) = \mathcal{Y}\}}.
    \end{equation*}
    \item \textbf{Mistrust}: Proportion of test samples without prediction, \textit{i.e.}, all the classes are too strange compared to the unclassified sample, and the sample cannot be classified with the specified confidence in any of the available classes:
    \begin{equation*}
        Mist = \frac{1}{k} \sum_{i=1}^{k} \mathbf{1}_{\{\Gamma_i^{\epsilon}(x_i) = \emptyset \}} .
    \end{equation*}
\end{itemize}
The performance assessment of classical single-predictions was conducted by the following well-known performance metrics: accuracy, precision and recall.  \textit{Per class} precision, recall and macro-F1 were also reported (see \textit{Supplementary Material}). \\

\subsection{Consistency evaluation}

A feature selection method should converge to similar subsets of features under random splits of a dataset. Ideally, the most informative features will be the ones selected more frequently.
A standard measure to assess the consistency between two sets of features with the same number of elements is the Jaccard index \cite{armañanzas_2011}.
The Jaccard index can be easily modified to consider differences between more than two sets. Let be \(S_1, ..., S_n\) subsets of features, such as \(|S_1| = ... = |S_n|\). The multi-set Jaccard consistency index is defined as
\begin{equation}
    \label{jac_2}
     I_{J} = \frac{|S_1\cap ... \cap S_n|}{|S_1\cup ... \cup S_n|}.
\end{equation}
We propose an extension that is more suitable than (\ref{jac_2}) to compare ensembles of \(n\) sets of features. Let consider \(|S_1| = ... = |S_n| \)  and \(K = (n/2 + 1,..., n)\), the new index \(I_W\) is defined as 
\begin{equation}
    \label{new_consistency}
    I_{W} =  \sum_{j \in K} \omega_j P_{j}, \;\; \text{such that} \;\; \omega_j = \frac{j}{\sum_{k \in K}k},
\end{equation}
where $P_{j} $ is the fraction of features that are common at least in \(j \in K\) subsets of features. 
This adaptation can be seen as a weighted Jaccard index. Unlike the expression in (\ref{jac_2}), the new index does not require that all generated subsets have a common set of features. A feature must be only present in at least \(n/2 +1\) of the subsets. The consistency measured by index \(I_W\) increases if a feature becomes more prevalent across diverse subsets. The proposed index ranges between 0 and 1, with higher values indicating more similarity between subsets, and it was conceived to detect the consistent subsets that the Jaccard index cannot detect.

\subsection{Experimental design}

Although our feature selection proposal is based on conformal prediction notions, if the inductive approach is followed, then our method can be described as conformal-agnostic. This implies that, once the feature selection method identifies the optimal subset of features, it is possible to implement both a traditional machine learning workflow, which provides point predictions, or a conformal prediction-based pipeline, which offers prediction intervals. We followed the inductive inference scheme to also optimize computational cost. The conducted performance comparison between CRFE and RFE methods was evaluated by the performance of the selected subsets of features in both the conformal prediction and traditional single prediction frameworks. When conformal prediction was used, we fixed the confidence level \(1 - \epsilon\) at 0.9. Test splits always included 25\% of the original dataset and the remaining samples were equally split into training and calibration sets. The parametric classifier chosen was a SVM with linear kernel, implemented through the \textit{Scikit-learn} library \cite{scikit-learn}.
The comparison scheme between both methods followed the next steps; (i) a random seed was established; (ii) both feature selection methods were run, selecting subsets of features for all possible sizes; and, (iii) predictions were inferred by both conformal and classical classifiers using the subsets of selected features. To carry out a fair, unbiased, and statistically robust comparison, the scheme was repeated 20 times by shuffling the data, and the results were averaged across all iterations. In each iteration, all the random numbers were generated and fixed at the beginning, always preserving the same split between test, calibration, and training sets for both selection methods.

\section{Results and discussion}\label{chap5}

\subsection{Performance analysis}\label{subsec:performance}

\begin{figure*}[!htb]
\centering
\subfloat[Synthetic dataset.]{\includegraphics[width=0.61\textwidth]{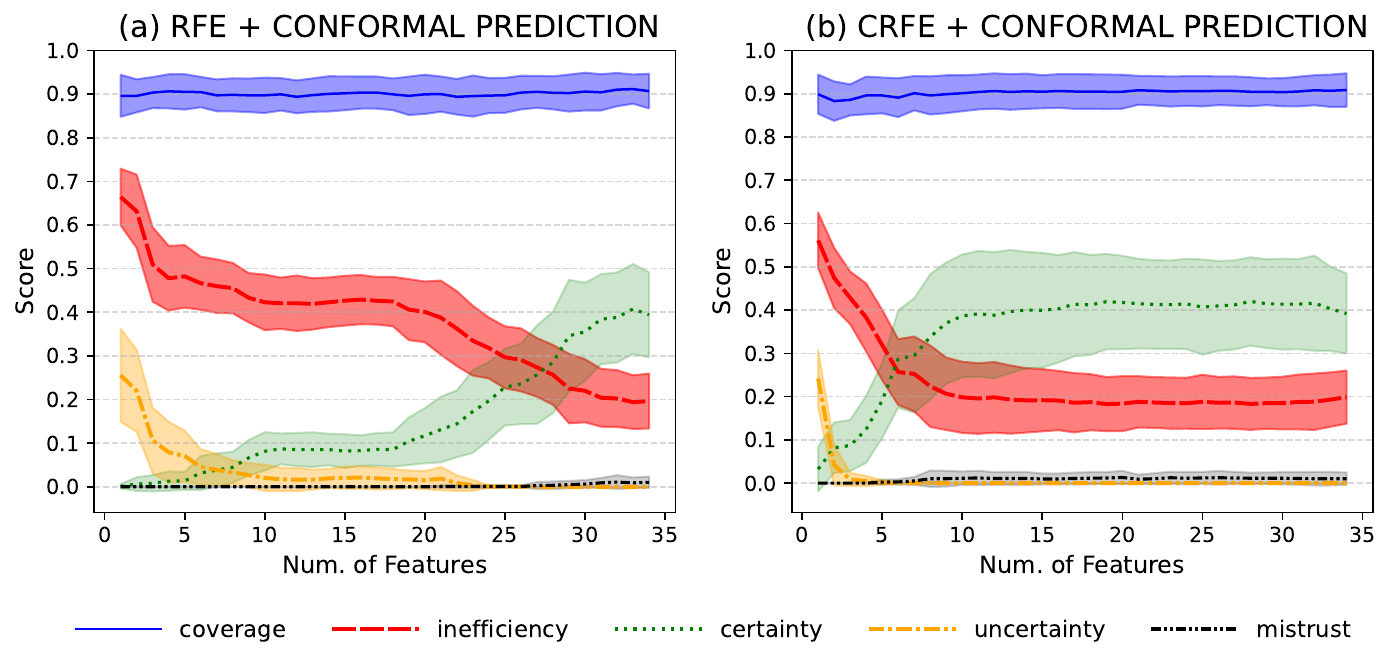}%
\label{fig_30}}
\hfil
\subfloat[Coronary artery dataset.]{\includegraphics[width=0.61\textwidth]{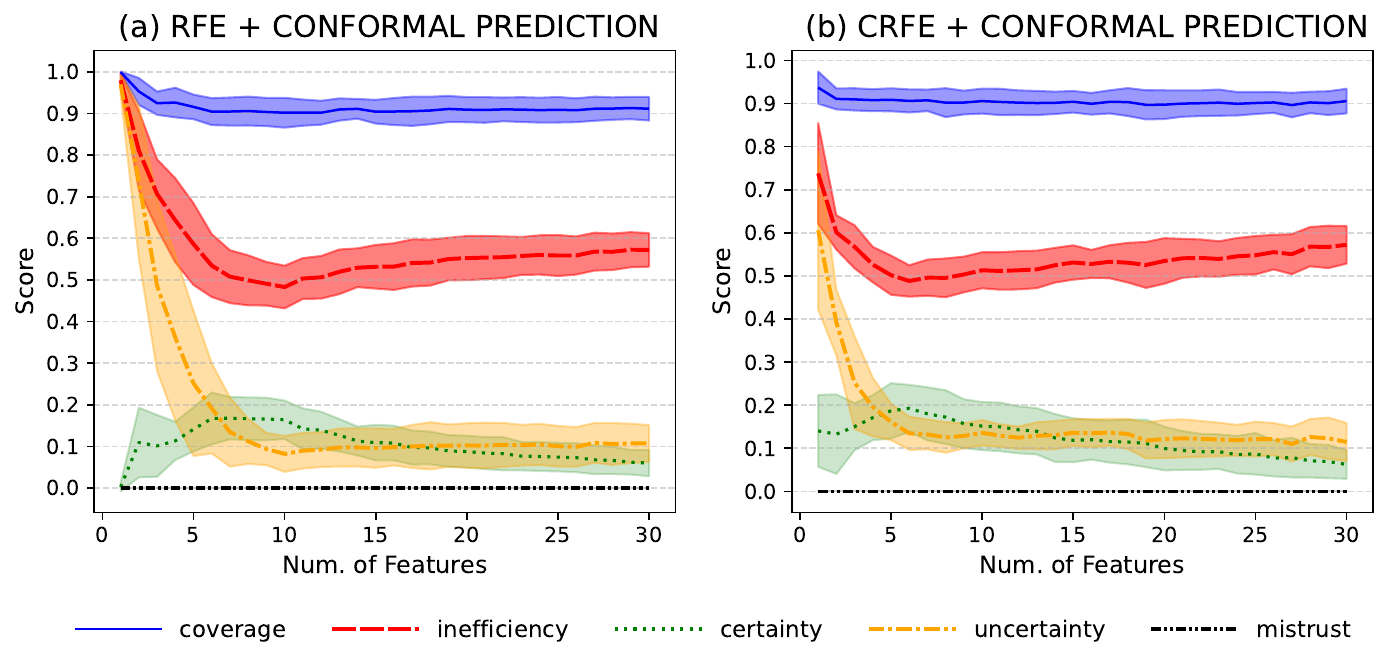}%
\label{fig_31}}
\hfil
\subfloat[Dermatology dataset.]{\includegraphics[width=0.61\textwidth]{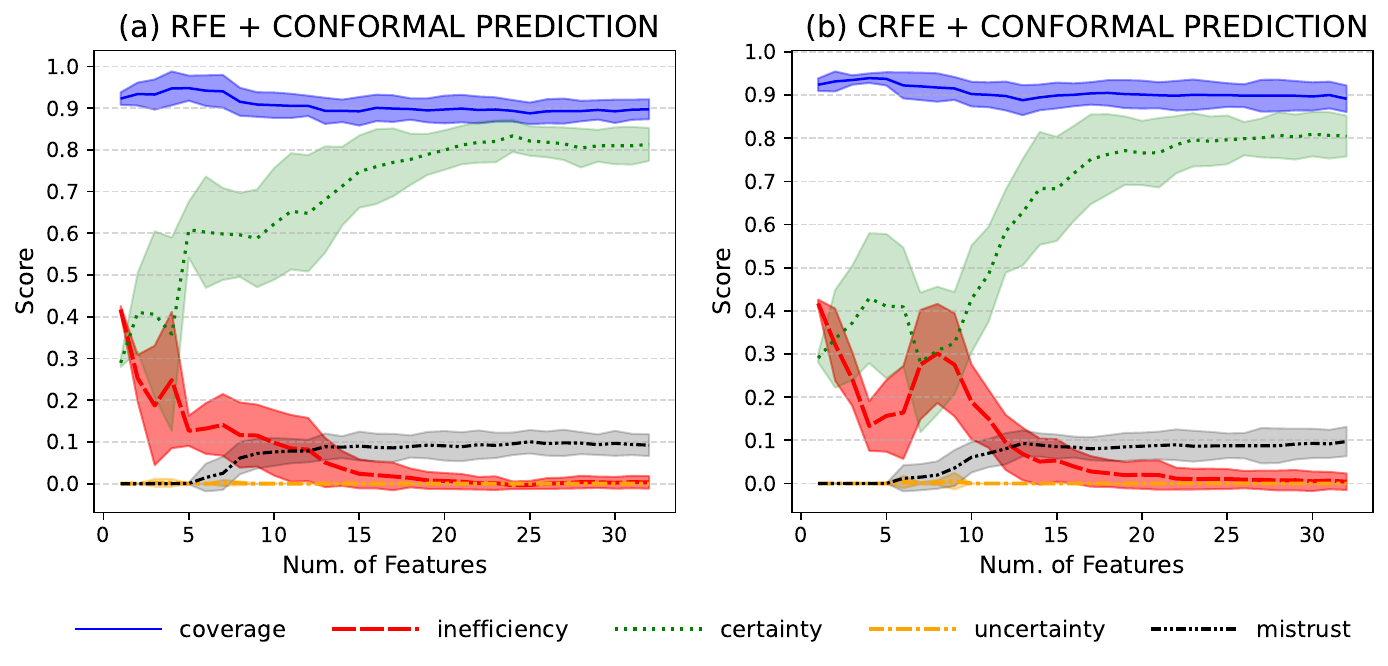}%
\label{fig_32}}
\hfil
\subfloat[Myocardial infraction dataset.]{\includegraphics[width=0.61\textwidth]{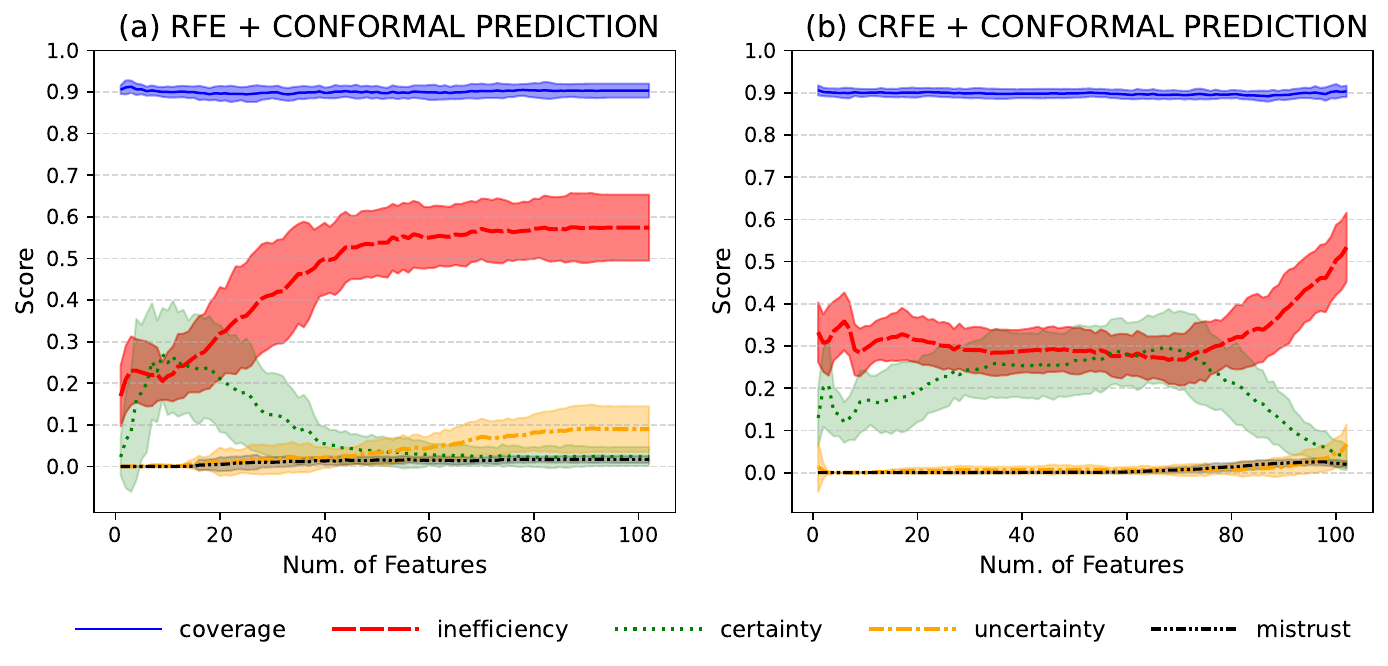}%
\label{fig_33}}
\caption{Set prediction performance metrics. The results are averaged over 20 train-test iterations for each feature selection method. Standard deviation is provided as upper and lower intervals. Plots a-(a), b-(a), c-(a), and d-(a) show results by the RFE method, whereas Plots a-(b), b-(b), c-(b), and d-(b) presents results by CRFE. }
\label{fig_sim}
\end{figure*}

The performance comparison for set predictions between RFE and CRFE methods is presented in Figures \ref{fig_30}, \ref{fig_31}, \ref{fig_32}, and \ref{fig_33} for the synthetic, coronary artery, dermatology, and myocardial infraction datasets, respectively.
We observed how CRFE outperformed RFE in the synthetic and the myocardial infraction datasets. Similar performance for both methods was observed in the dermatology and the myocardial infraction datasets.\\

On the synthetic dataset, CRFE outperformed RFE performance scores. The inefficiency metric, \textit{i.e.}, the average length of the prediction sets, is clearly smaller when the CRFE method was applied than in the case of the RFE. Moreover, certainty and uncertainty scores in Figure \ref{fig_30}-(b) support the fact that CRFE was able to find the features that minimize the strangeness between samples, without losing classification performance. On the other hand, Figure \ref{fig_30}-(a) shows how the performance of the scores start worsening faster when RFE was applied. In this dataset the number of features that are known to be the informative was configured to 10. On average, at least 75\% of the informative features were constantly present when CRFE method was used, whereas the RFE captured less than 14\% of informative features across all runs.\\

The comparison performance on the coronary artery disease dataset showed a similar trend for both feature selection methods. The uncertainty and inefficiency scores displayed a pronounced growth in Figure \ref{fig_31}-(a) for subsets of features with less than 9 features, whereas the same behaviour was observed in Figure \ref{fig_31}-(b) for subsets of features with less than 6 features. Therefore, the optimal size of features proposed by RFE had a cardinality of 9 to 10, whereas the optimal subset proposed by CRFE had 5 to 6 features.\\

Results for the dermatology dataset showed similar performance by both feature selection methods, see Figures \ref{fig_32}-(a) and \ref{fig_32}-(b). Optimal cardinalities were found for subsets in the range of 15 to 20. However, CRFE found a second optimal subset lowering this range just to 5 features. This second optimal set scored a mistrust of 0. Finally, on the myocardial infraction dataset, a clear difference between both methods was observed. Figure \ref{fig_33}-(a) shows an optimal point for the RFE performance close to subsets of size 10, with null uncertainty value at that cardinality.\\

\begin{figure*}[!htb]
\centering
\subfloat[RFE.]{\includegraphics[width=0.7\textwidth]{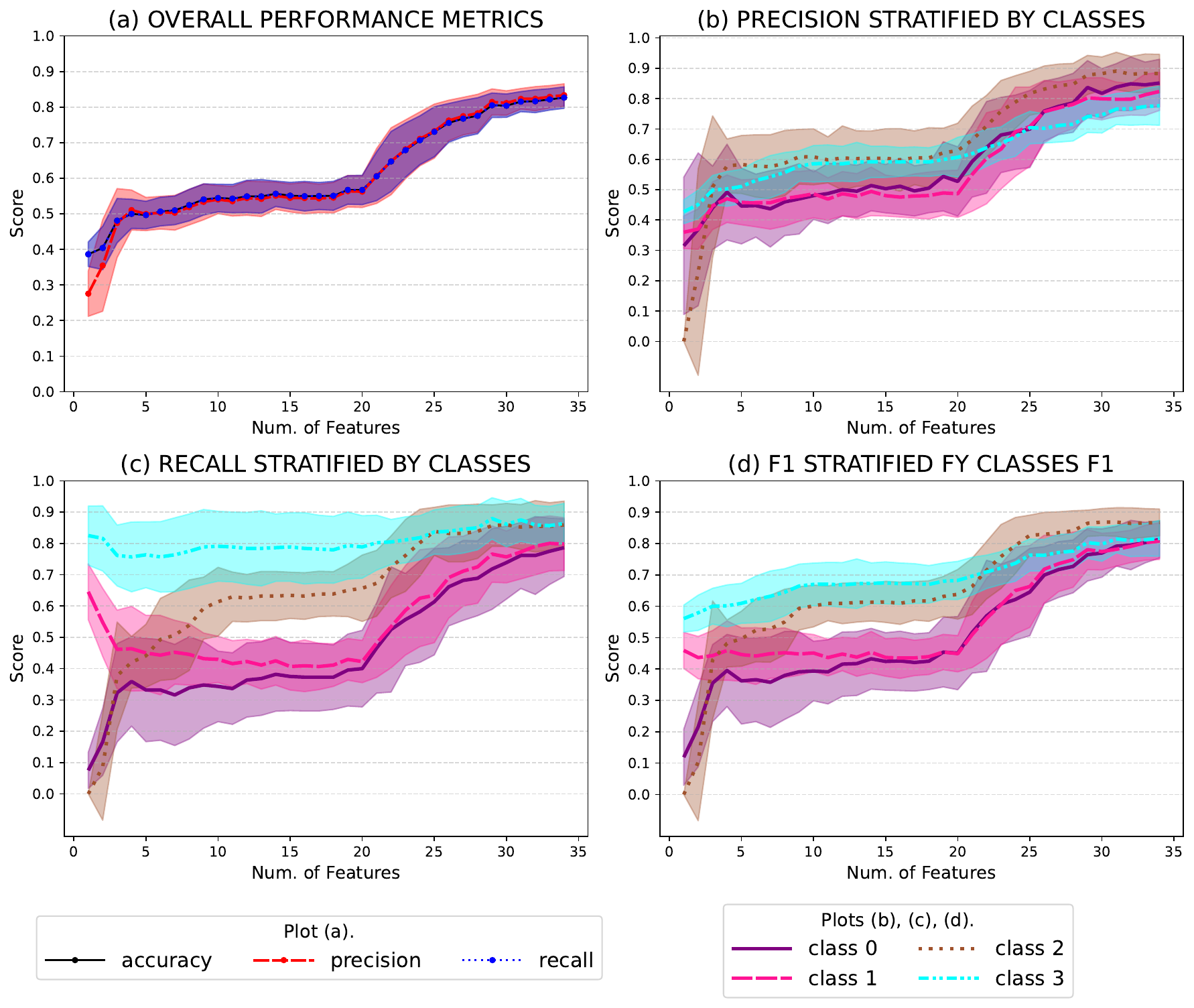}%
\label{fig_cla_a}}
\hfil
\subfloat[CRFE.]{\includegraphics[width=0.7\textwidth]{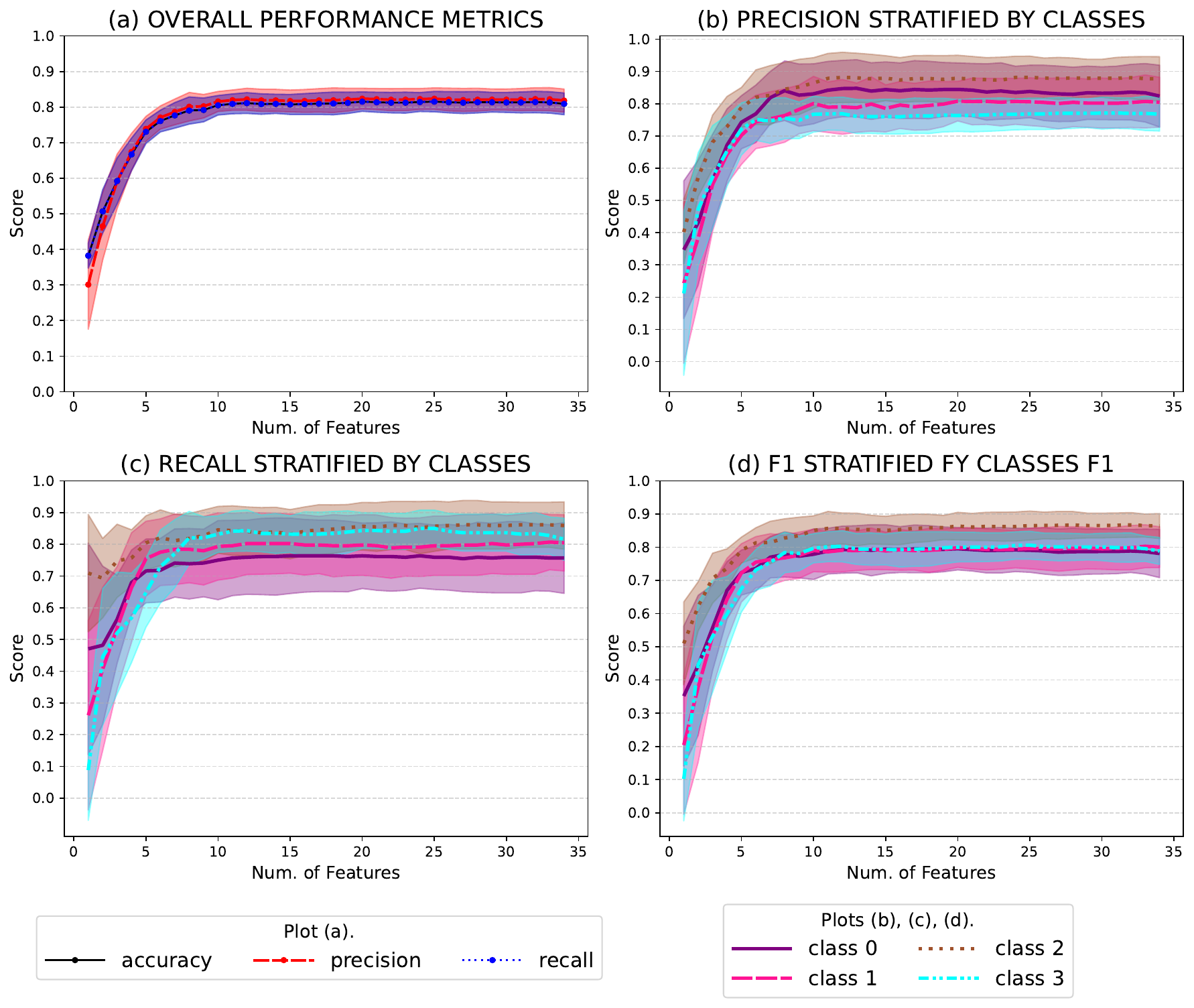}%
\label{fig_cla_b}}
\caption{Single-prediction performance metrics for the synthetic dataset. Standards deviations are provided as upper and lower intervals. Plots a-(a) and b-(a) show accuracy, precision, and recall performance metrics achieved by subsets of features selected by RFE and CRFE, respectively. Plots a-(b),(c),(d) and b-(b),(c),(d) show precision, recall and F1 score performance by class achieved by RFE and CRFE, respectively.  }
\label{fig_class_syn}
\end{figure*}

The single-prediction performance metrics also showed the advantage of CRFE when classical machine learning prediction was implemented. Figure \ref{fig_class_syn} shows the results for single-prediction performance metrics both for RFE and CRFE feature selection methods on the synthetic dataset. CRFE was able to preserve overall scores removing features until subset sizes of 10, which corresponds with the subset size of the known predictive features. Stratified by class results showed a similar behavior. 
Results for the remaining datasets are provided in the \textit{Supplementary Material}.

\subsection{Consistency analysis}\label{subsec:Consistency}

\begin{figure*}[!htb]
\centering
\subfloat[Synthetic dataset.]{\includegraphics[width=0.61\textwidth]{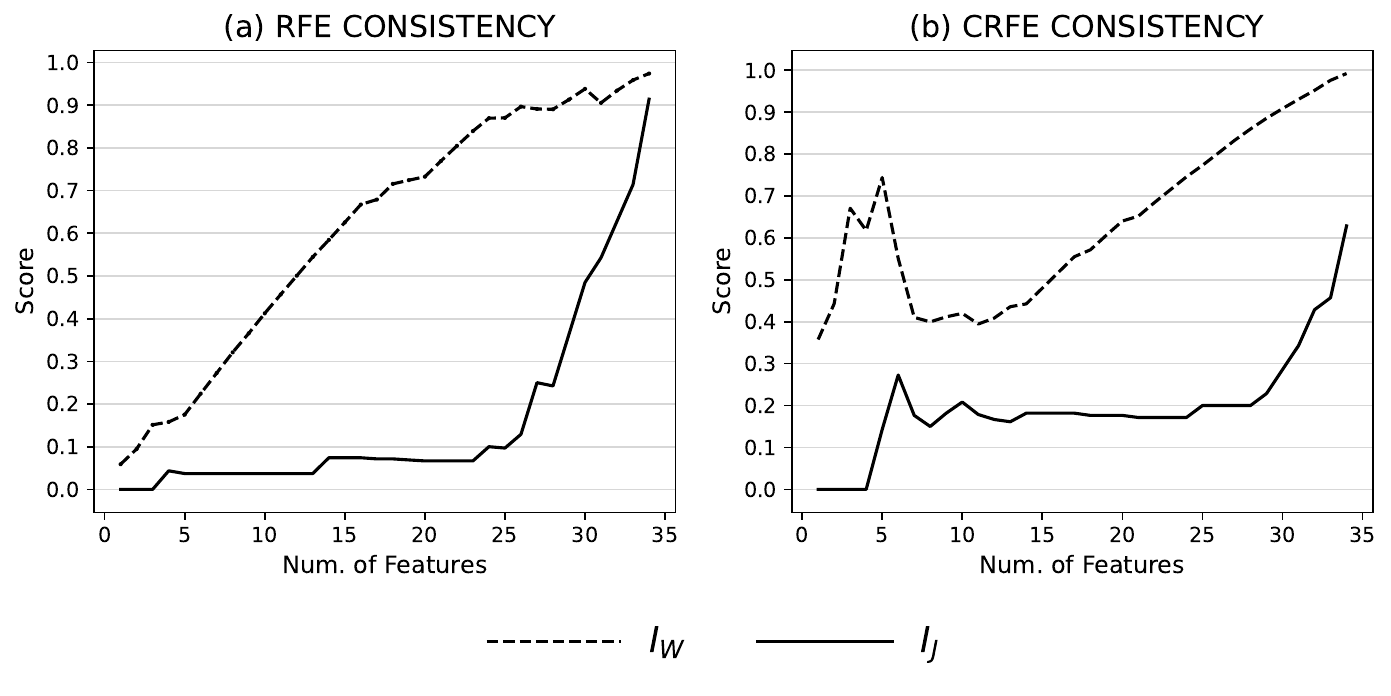}%
\label{FEAT_30}}
\hfil
\subfloat[Coronary artery dataset.]{\includegraphics[width=0.61\textwidth]{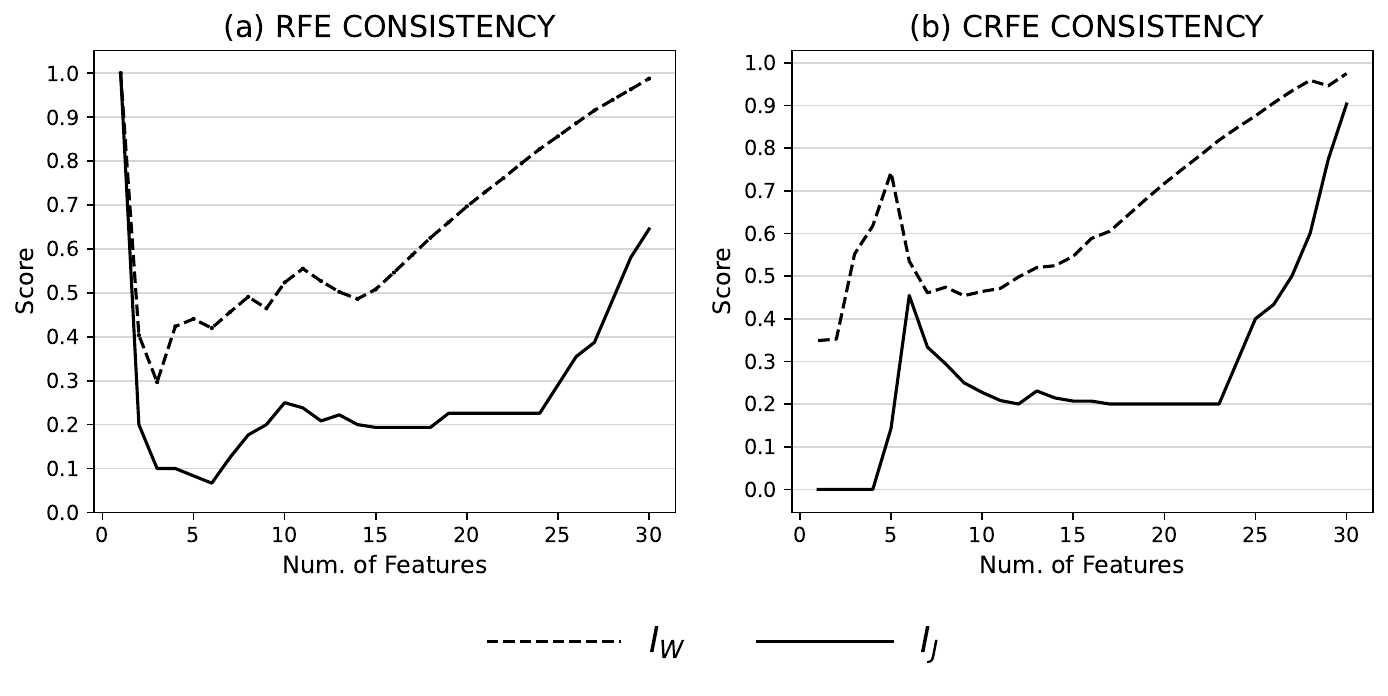}%
\label{FEAT_31}}
\hfil
\subfloat[Dermatology dataset.]{\includegraphics[width=0.61\textwidth]{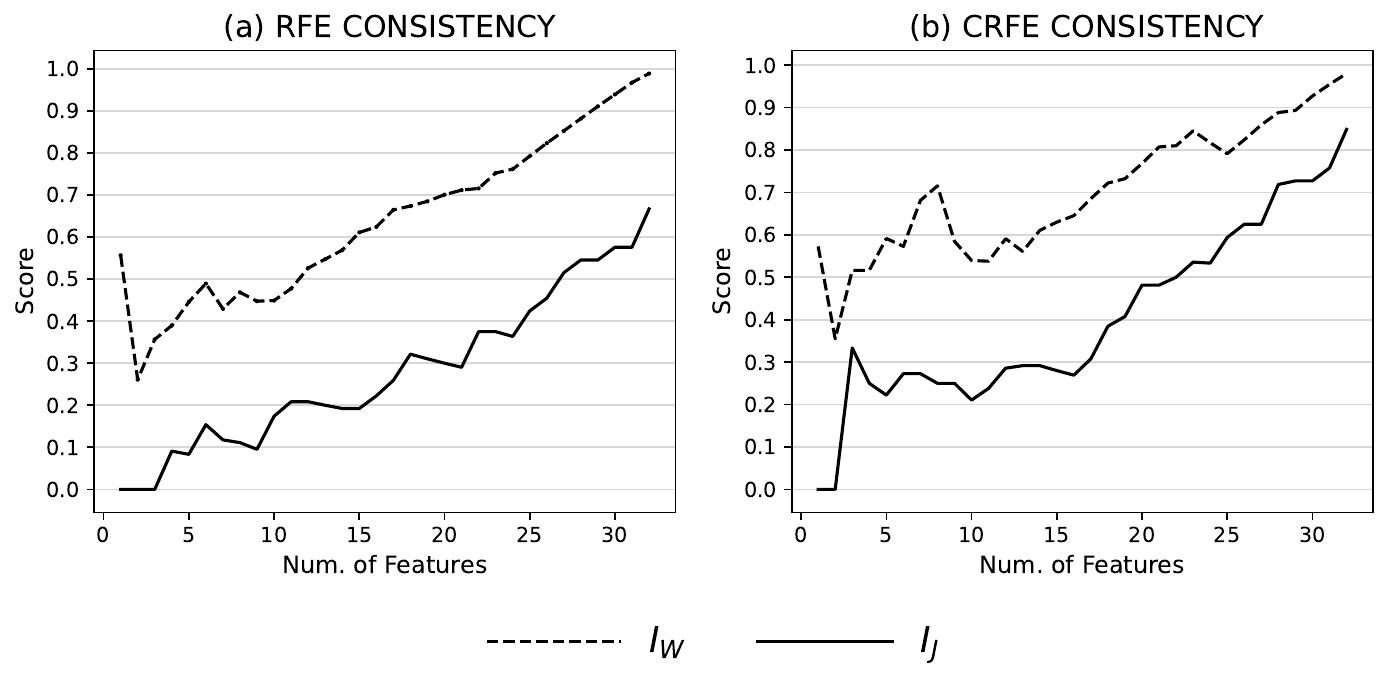}%
\label{FEAT_32}}
\hfil
\subfloat[Myocardial infraction dataset.]{\includegraphics[width=0.61\textwidth]{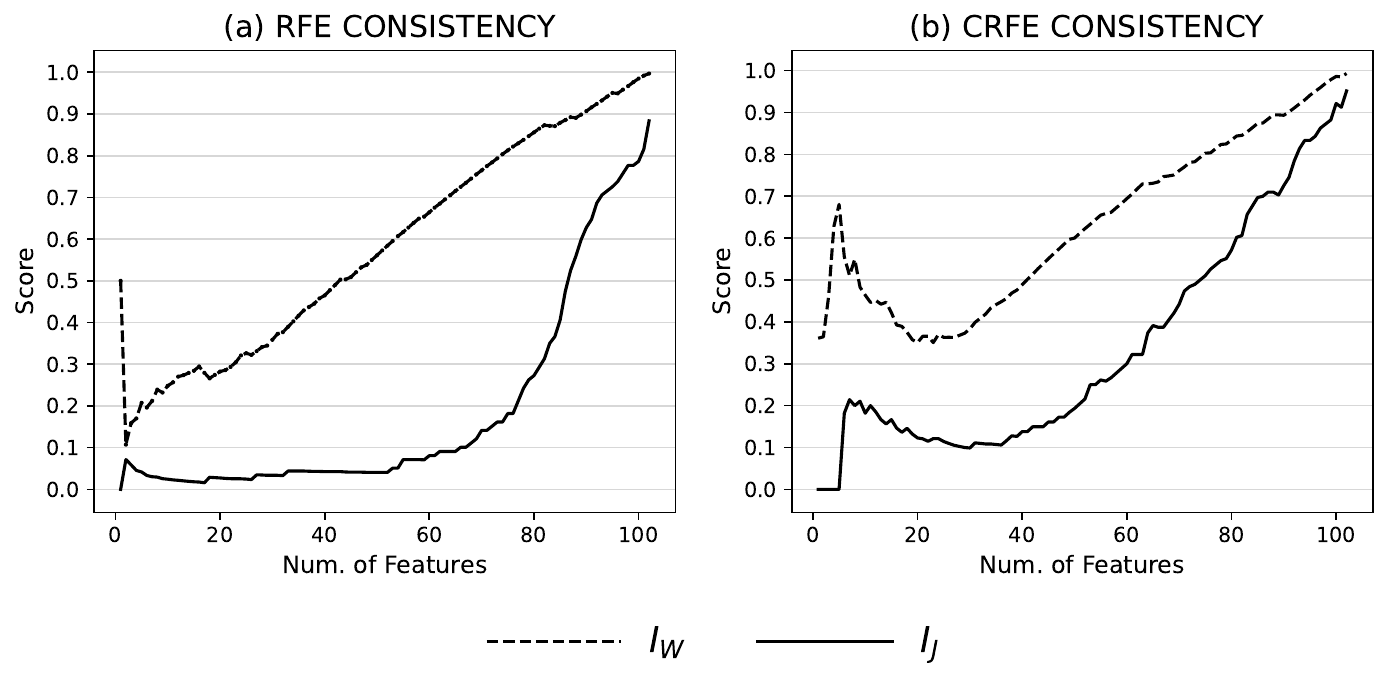}%
\label{FEAT_33}}
\caption{Consistency analysis across the 20 iterations. Plots a-(a), b-(a), c-(a) and d-(a) show consistency results for RFE, whereas Plots a-(b), b-(b), c-(b) and d-(b) present CRFE consistency results. The Jaccard \(I_J\), and the new proposed consistency index \(I_W\) are shown. }
\label{consistency_analysis}
\end{figure*}

We also analyzed how consistent the features selected by CRFE were in comparison to RFE.
Figure \ref{consistency_analysis} presents the results obtained by the Jaccard and the novel consistency indexes described in Expressions (\ref{jac_2}) and (\ref{new_consistency}), respectively. These indexes quantify the level of consistency between the subsets of features selected by both RFE and CRFE for each of the 20 iterations.\\ 

Note how the feature subsets selected by CRFE were equally or more consistent than those selected by RFE. The peaks of consistency, especially in Figures \ref{FEAT_30}-(b), \ref{FEAT_31}-(b), were observed around the optimal subset sizes predicted in Figure \ref{fig_sim} for CRFE. This suggests that CRFE was able to, independently of the data split, find the subset of features which reports better performance metrics. We highlight how the newly proposed consistency index in Expression (\ref{new_consistency}) was able to reflect the peaks in Figures \ref{FEAT_30}-(b), \ref{FEAT_32}-(b), \ref{FEAT_33}-(a), and \ref{FEAT_33}-(b), in contrast to the Jaccard index. These peaks contributed to detect hidden consistent subsets. Moreover, the new index reflected consistency at the last stages of the recursive elimination process while Jaccard ignored those sizes. In Figures \ref{FEAT_30}-(a) and \ref{FEAT_33}-(a) there are noticeable differences between those consistency measures. Those values may indicate that there was a considerable number of features that, although not consistently selected in all the 20 iterations, were frequently chosen by RFE. For coronary artery and dermatology datasets, both showed a similar behavior for larger feature subsets. For smaller subsets, CRFE showed more consistent selections than RFE.\\

We also analyzed the differences between the features selected by both methods when the random seeds were the same, \textit{i.e.}, similarity. We compared the subsets of features selected by both methods with (i) the same cardinality and (ii) selected in the same iteration, see results in Suplementary Figure S4.
The feature subsets selected had greater similarity for the coronary artery and dermatology datasets. The Kuncheva index defined in Supplementary Equation S6.1 revealed that these coincidences were robust against random coincidence. On the other hand, synthetic and myocardial infraction datasets revealed poor agreement on the selected features. For these two datasets the similarity values suggested that the agreement by chance was relevant.

\subsection{ \(\beta\)-based stopping criterion}\label{stopp_cri_ev}

The \(\beta\)-based stopping criterion introduced in Section \ref{Sub_stopp} was tested in all four datasets. We compared the results of the new criterion versus the accuracy-based RFE stopping criterion. RFE accuracy-based rule is a well established method implemented through a cross validation pipeline. We tested the RFE-accuracy and the \(\beta\)-based stopping criterion by running 50 different random splits of the datasets. We fixed the excess \(\sigma\), \textit{i.e.}, the number of times the second derivative must exceed the standard deviation to stop the recursive method, to 5 in Algorithm \ref{alg:alg1}. Results are presented in Table \ref{stopping_table}. The distribution of selected features by the \(\beta\)-based stopping criteria are shown in Figure \ref{fig_hist}. The correspondent distribution of subset sizes are included in Supplementary Figure S5.\\ 

\begin{table*}[h]
\label{stopping_table}
\centering
\resizebox{0.99\textwidth}{!}{
\begin{tabular}{lrr}
\hline
\hline
\textbf{Dataset} & \textbf{RFE} &  \textbf{C-RFE}   \\

& \textbf{Average size (Inefficiency, Certainty)}  & \textbf{Average size (Inefficiency, Certainty)}    \\
\hline
 \textbf{Synthetic} & 34 (0.19, 0.39)  & 11 (0.19, 0.39)  \\ 

 \textbf{Coronary artery} &  16 (0.53, 0.10)  & 8 (0.49, 0.17)  \\ 

 \textbf{Dermatology} & 31 (0.04, 0.81)  & 16 (0.04, 0.72)   \\
 \textbf{Myocardial infarction} & 1 (0.16, 0.02)  & 17 (0.32, 0.17) \\
 \hline
 \hline
\end{tabular}
}
\caption{Average size for subsets of features selected by both stopping criteria: RFECV for RFE and \(\beta\)-based for CRFE. Results inside parentheses are the averaged inefficiency and certainty scores for the selected subset size.}
\end{table*}

In the synthetic dataset, the optimal size found by the RFE-accuracy criterion significantly differed from the most frequent subset size proposed by the \(\beta\)-based criteria. Results in Figure \ref{fig_hist}-(a) show that the distribution of the most frequent selected features was consistent with the known informative features.
For the coronary artery dataset, 9 and 10 were the subset sizes most commonly observed when the \(\beta\)-based criterion was used, closely followed by sizes of 6 and 7 elements, see Supplementary Figure S5-(b). Figure \ref{fig_hist}-(b) shows that there were 6 features consistently selected at least in 80\% of the experiments. These findings are in line with the results in Figures \ref{fig_31} and \ref{FEAT_31}. 
In the dermatology dataset, the RFE method discarded only one feature, whereas the new criterion frequently stopped at three different subset sizes: 13, 20, and 25. Two of the optimal subsets with 20 and 25 features by the \(\beta\)-based criterion were consistent with the optimal size shown in Figure \ref{fig_32}-(b). The RFE-accuracy based criterion was far from any optimal subset size from results shown in Figure \ref{fig_32}-(a).
In the myocardial infarction dataset, both stopping criteria resulted in subset sizes that deviated significantly from the optimal subsets detected in Figure \ref{fig_33}. However, the inefficiency and certainty metrics in Table \ref{stopping_table} were better when the new stopping criterion was used.

The \(\beta\)-based automatic stopping criteria was designed to determine the optimal subset of features without computing performance scores for all possible feature cardinalities or combinations. We observed that variability in the data can slightly influence its performance. However, the frequency of features included in the subsets selected by the new criterion showed that this method was able to identify the most significant features to effectively reduce non-conformity. 

\begin{figure*}
    \centering
    \includegraphics[width=0.95\textwidth]{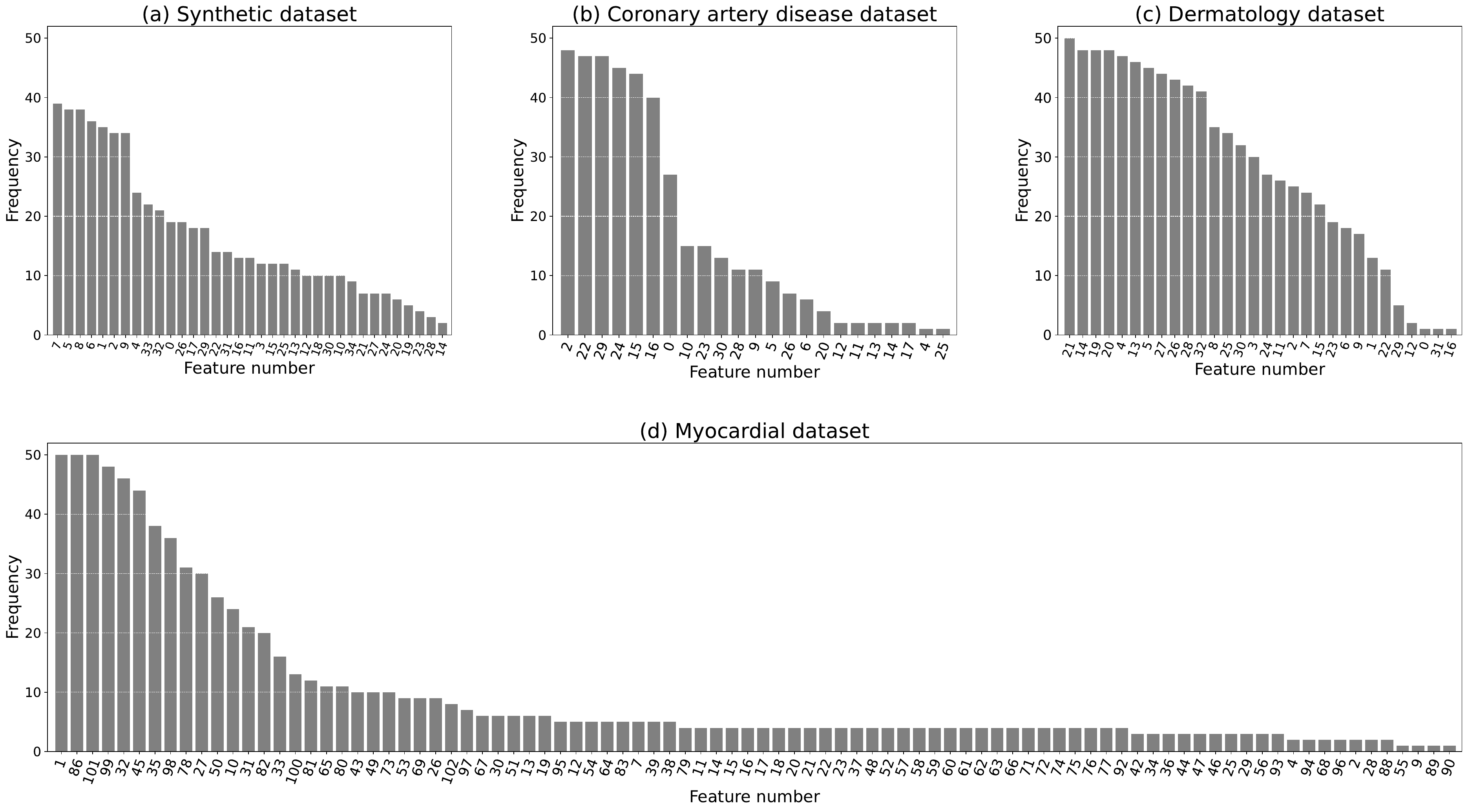}
    \caption{The frequency with which each feature was included in the optimal subset of features selected by CRFE using the \(\beta\)-based stopping criterion. Note the maximum corresponds to 50 independent runs. The percentages of features always discarded (not shown) were 0\%, 28.2\%, 11.7\%, and  13.5\% of the total sets for synthetic, coronary artery disease, dermatology, and myocardial datasets, respectively. }
\label{fig_hist}
\end{figure*}

\section{Conclusions}\label{chap6}

Conformal prediction stands out as one of the most effective methods for uncertainty quantification due to its robust theoretical guarantees.
In this study, we introduce an alternative to the classical Recursive Feature Elimination method but taking advantage of the conformal prediction framework. The proposed method, named Conformal Recursive Feature Elimination (CRFE), is a feature selection technique that builds upon the SMFS algorithm \cite{Bellotti_et_al_2006}, but extended to recursively remove features. We demonstrate that, by removing a feature, the variation in the non-conformity measure is equivalent to eliminating the non-conformity associated with that particular feature. Additionally, we introduce a novel consistency index and an automatic stopping criterion based on the non-conformity associated with the features. To evaluate the effectiveness of CRFE, we compared its performance against RFE on four multiclass datasets. The results indicate several benefits of using CRFE when conformal prediction is applied, as well as in classical performance metrics when no conformal prediction is done. The consistency tests based on stability indexes reported that CRFE achieved at least the same level of consistency as RFE. 
The proposed automatic stopping criterion for CRFE is based on the non-conformity values of each feature and outperformed the accuracy-based RFE stopping criterion. However, the feature selection method proposed in this work depends on the linear separability condition when the selection is performed, \textit{i.e.}, CRFE relies on computing separation hyperplanes between classes. Future developments will investigate how to expand to nonlinear classifiers. We also plan to explore the class-conditional conformal prediction as a mechanism to better adapt to imbalanced problems. Finally, an open source library with the implementation is released.

\section*{Acknowledgments}
This work was supported by the Gobierno de Navarra through the ANDIA 2021 program (grant no. 0011-3947-2021-000023) and the ERA PerMed JTC2022 PORTRAIT project (grant no. 0011-2750-2022-000000).

\section*{Repository}
An open source library that implements CRFE, including the datasets used in this work, the post processed data, and the code can be found at \url{https://github.com/digital-medicine-research-group-UNAV/CRFE}.

\appendix
\section{Supplementary Material}

\subsection{Introduction}

The present document provides \textit{Supplementary Material} for the paper entitles \textit{Conformal Recursive Feature Elimination}. The results presented here were derived from the same datasets as shown in the main document. Single-prediction performance metrics, \textit{i.e.}, accuracy, recall, and precision, as well as \textit{Per class} precision, recall and F1 metrics were calculated in supplementary Sections \ref{Sec:Cor}, \ref{Sec:Der}, and \ref{Sec:Myo} for the last three described datasets. Precision is defined as the the proportion of predicted positives that are truly positive. Recall, also known as sensitivity, provides the proportion of positives that are correctly classified. The F-1 score represents the harmonic mean between precision and recall. The plots supporting the similarity study are exposed in supplementary Section \ref{Sec:Sim}, and the additional material related to the study of the \(\beta\)-based stopping criterion is provided in supplementary Section \ref{Sec:Stop}.

\subsection{Artificial dataset} \label{Sec:Art}

The synthetic dataset was generated using the \textit{make\_classification()} method from the \textit{scikit-learn} library. The specific parameters used are detailed in Supplementary Table \ref{supp_1}.
\begin{table}[H]
\centering
\begin{tabular}{lrlr}
\hline
\hline
\textbf{n\_samples} & 350 &  \textbf{n\_redundant}  & 1  \\
\textbf{n\_informative}  & 10   &  \textbf{n\_clusters\_per\_class} &  1 \\
\textbf{n\_classes}  & 4   &  \textbf{class\_sep} & 1.5  \\
\textbf{random\_state}  & 12345  &  \textbf{flip\_y} & 0.05  \\
 \hline
 \hline
\end{tabular}
\caption{Parameters used to generate the synthetic dataset.}
\label{supp_1}
\end{table}

\subsection{Coronary artery disease dataset}\label{Sec:Cor}

The coronary artery disease dataset \cite{UCI} comprises data collected from Cleveland (303), Hungary (294), Switzerland (123) and Long Beach VA (200). The original study  \cite{Detrano_1989} warned about a potential bias in the test groups because the noninvasive samples were not withheld from the treating physician. To classify diseases, a cardiologist diagnosed samples only based on the angiogram results. Accordingly to his criteria, a coronary artery was significantly diseased if the luminal diameter reduction exceeded 50\%.\\

\begin{figure}[!htb]
\centering
\subfloat[RFE.]{\includegraphics[width=0.7\textwidth]{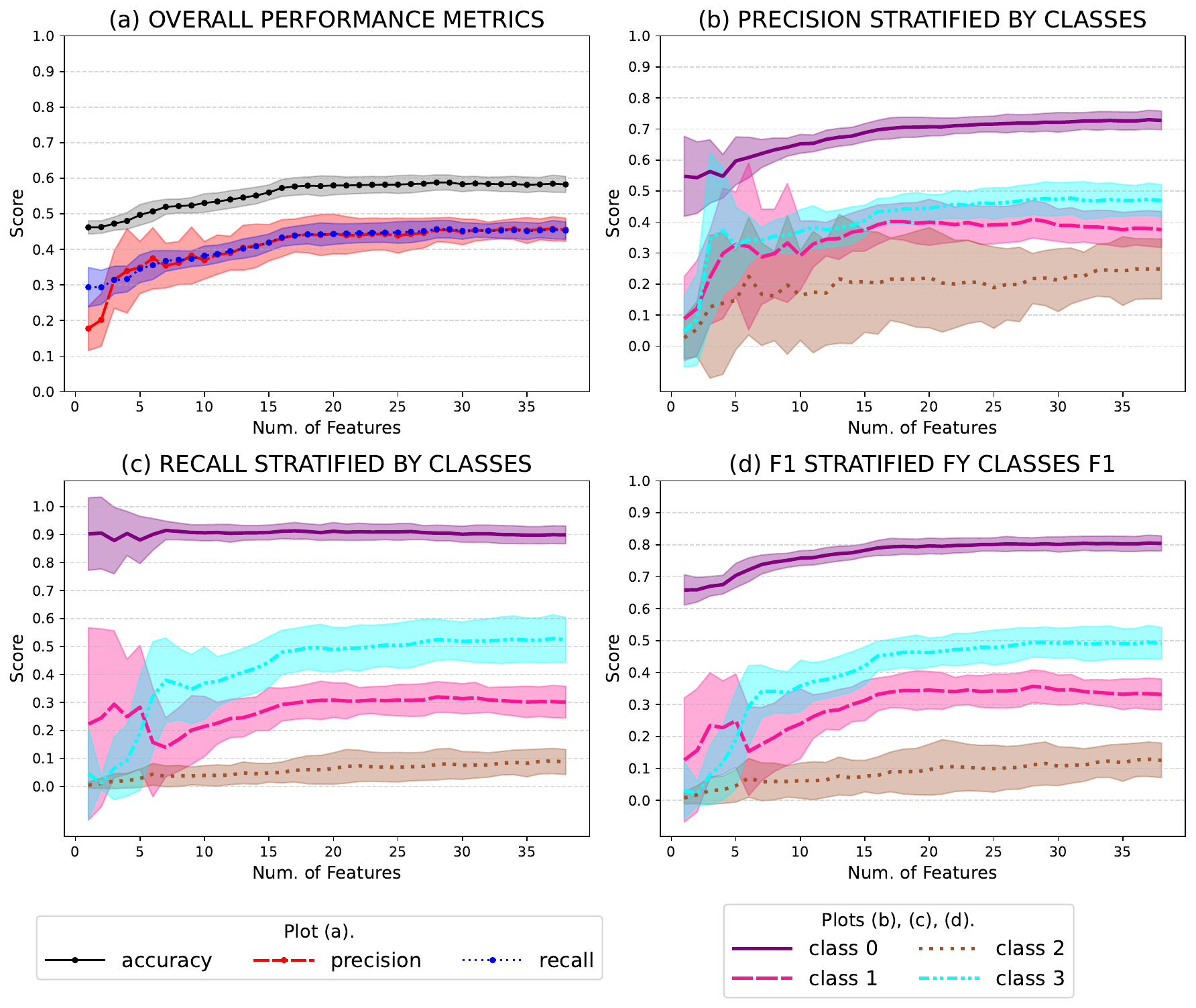}%
\label{fig_1_a}}
\hfil
\subfloat[CRFE.]{\includegraphics[width=0.7\textwidth]{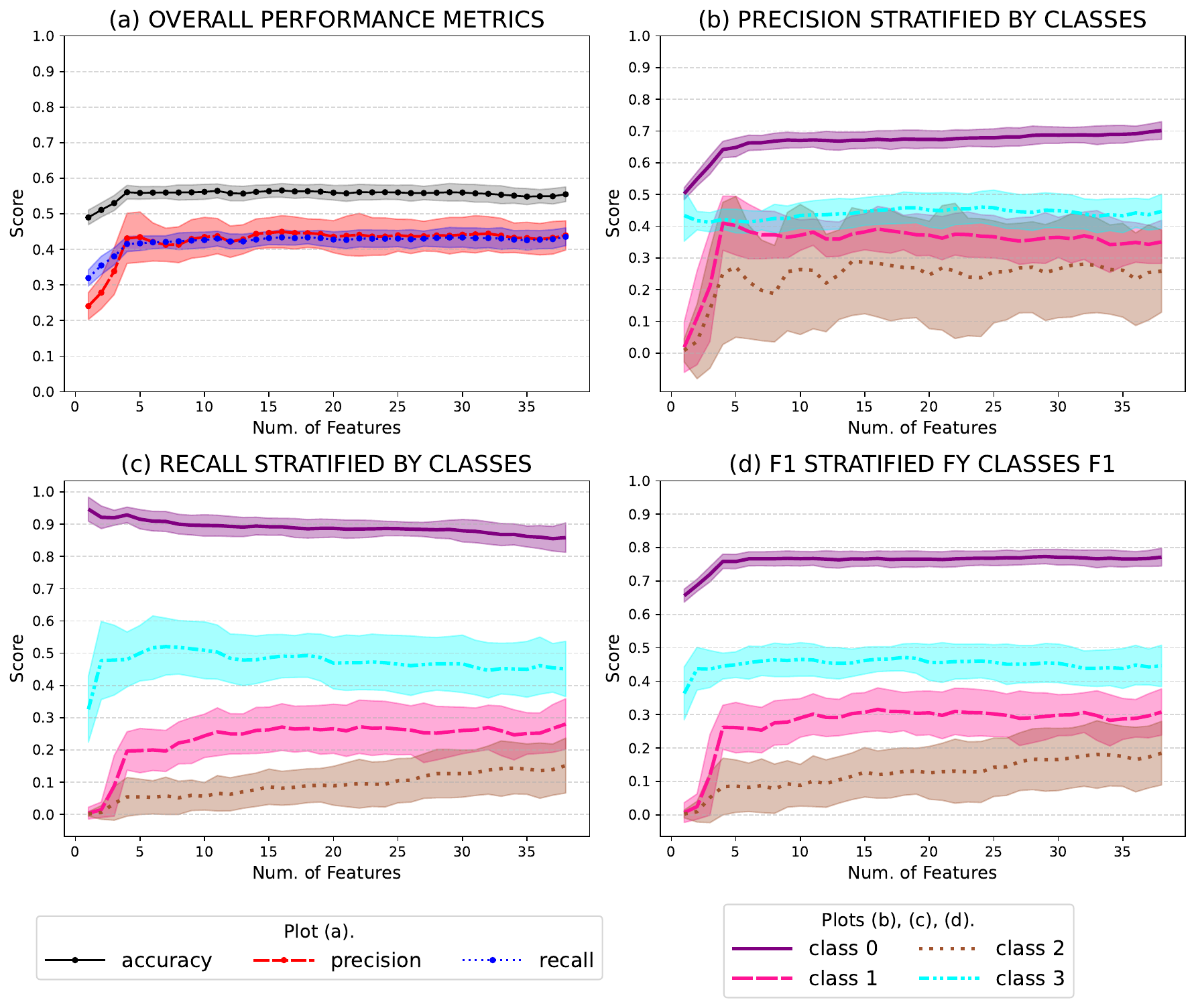}%
\label{fig_1_b}}
\caption{Single-prediction performance metrics for the coronary artery disease dataset. Standard deviations were provided as upper and lower intervals. Plots a-(a) and b-(a) show accuracy, precision, and recall performance metrics achieved by subsets of features selected by RFE and CRFE respectively. Plots a-(b),(c),(d) and b-(b),(c),(d) show precision, recall, and F1 score performance by class achieved by RFE and CRFE, respectively.  }
\label{Figure_1}
\end{figure}

The classes in the dataset are; \textit{class 0} stands for no disease and has 404 samples of the total, \textit{i.e.} 44.93\%, \textit{class 1} stands for patients that have at least one diseased artery and is composed of 191 samples (21.24\%), \textit{class 2} stands for those patients that have a single-vessel disease and is composed of 130 samples (14.46\%), \textit{class 3} stands for those that has a double-vessel disease and is composed of 132 samples (14.68\%), and \textit{class 4} stands for those that has a triple-vessel disease and is composed of 42 samples (4,67\%). The original database comprised 75 features, but \textit{name, SS number, medical proofs dates} and \textit{patient´s number} were excluded. Any feature that had more than 25\% of missing values was also removed. The remaining features are listed and numbered in Supplementary Table S\ref{supp_2}. We encourage the reader to see the publicly available repository for detailed information of the specific features considered. Missing data were imputed by a K-nn algorithm with 5 neighbors. Samples were standardized to avoid scale biases.
\begin{table}[H]
\centering
\resizebox{1\textwidth}{!}{
\begin{tabular}{llllll}
\hline
\hline
\textbf{Number} & \textbf{Feature label} &  \textbf{Number} & \textbf{Feature label} &\textbf{Number} & \textbf{Feature label}  \\
   \hline
 0 & \textbf{age} & 13 & \textbf{proto} & 26 & \textbf{rldv5e}  \\ 
 1 & \textbf{sex} & 14 & \textbf{thaldur} & 27 & \textbf{lvx1}  \\ 
 2 & \textbf{cp} & 15 & \textbf{thaltime} & 28 & \textbf{lvx2}  \\ 
 3 & \textbf{trestbps} & 16 & \textbf{met} & 29 & \textbf{lvx3}  \\ 
 4 & \textbf{htn} & 17 & \textbf{thalach} & 30 & \textbf{lvx4}  \\ 
 5 & \textbf{chol} & 18 & \textbf{thalrest} & 31 & \textbf{lvf}  \\ 
 6 & \textbf{fbs} & 19 & \textbf{tpeakbp} &  &   \\ 
 7 & \textbf{restecg} & 20 & \textbf{tpeakbpd} & &   \\ 
 8 & \textbf{dig} & 21 & \textbf{dummy} &  & \\ 
 9 & \textbf{prop} & 22 & \textbf{trestbpd} &  &   \\ 
 10 & \textbf{nitr} & 23 & \textbf{exang} & &   \\ 
 11 & \textbf{pro} & 24 & \textbf{xhypo} &  &   \\ 
 12 & \textbf{diuretic} & 25 & \textbf{oldpeak} &  &   \\ 
 \hline
 \hline
\end{tabular}
}
\caption{Features in coronary artery disease dataset. Features considered after data pre-proccessing.}
\label{supp_2}
\end{table}
Supplementary Figures \ref{fig_1_a} and \ref{fig_1_b} support results presented in Section 5 of the main document.
Results presented in Figure 4 of the main document showed that features selected by the \(\beta\)-based stopping criteria more than 80\% of the cases were:  \textit{cp} which stands for chest pain type, \textit{thaltime} which stands for the time when ST measure depression was noted in the ECG, \textit{met} which is true or false if a threshold on Methabolic Equivalent (MET) is achieved while exercise testing, \textit{trestbpd} which stands for the resting blood pressure, \textit{xhypo} which stands for exercise-induced hypotension and \textit{lvx3} which was not explained either in the dataset documentation and the original work, but we decided to include in the dataset because we postulate that it stands for some type of pacing interval with the left ventricle, which is relevant medical information.

\subsection{Dermatology dataset}\label{Sec:Der}

The dermatology dataset includes a classification problem between 6 classes \cite{UCI}. These classes correspond with differential diagnoses of \textit{erythemato-squamous} disease: \textit{Psoriasis} class have 112 samples (30.60\%), \textit{Seborrheic Dermatitis} class 61 (16.67\%), \textit{Lichen Planus} class 72 (19.67\%), \textit{Pityriasis Rosea} class 49 (13.39\%), \textit{Chronic Dermatitis} class 52 (14.21\%), and \textit{Pityriasis Rubra Pilaris} class 20 (5.46\%). Firstly, patients were evaluated through 12 clinical features. Subsequently, 22 histopathological features were annotated from a biopsy. These features are listed in Table \ref{supp_3}. The original work \cite{Altay_1998} warned that, while some samples exhibited the typical histopathological features of the disease, others did not. The original work reported an accuracy score of 99.2\%, but no other performance was provided. In this work, both RFE and CRFE methods consistently maintained accuracy, precision, and recall scores above 95\% until the recursive elimination of features began to deteriorate the results. The optimal subset of features appears to be around 8 to 12 elements for RFE, see supplementary Figure \ref{fig_2_a}. The optimal subset size of features for CRFE, accordingly to results showed in supplementary Figure \ref{fig_2_b}, seems to range between 15 to 20 elements.
Figures \ref{fig_2_b}-(b), \ref{fig_2_b}-(c) and \ref{fig_2_b}-(d) showed that CRFE was able to provide the necessary information to narrow down the intervals of relevant features by class. On the other hand, the results presented by RFE in Figures \ref{fig_2_a}-(b), \ref{fig_2_a}-(c) and \ref{fig_2_a}-(d) indicated that the scores associated with each class start getting worse for same subset sizes. However, the scores provided by CRFE identified sets of features responsible for the deterioration of the class-performance scores.\\

The original work proposed the use of a genetic algorithm to determine the relevance of each feature. The most relevant features found by the original work were \textit{koebner} and \textit{inflammatory mononuclear  inflitrate} features, while the features \textit{acanthosis, follicular
horn plug, munro microabcess}, and \textit{age} were found to be the least relevant. Results presented in Figure 4 of the main document partially agrees with this. The least relevant features proposed by the original paper were also selected as the least relevant by CRFE. \textit{Koebner} was also selected as one of the most relevant. However, \textit{inflammatory mononuclear inflitrate} was almost irrelevant for CRFE.
\begin{table}[H]
\centering
\resizebox{0.99\textwidth}{!}{
\begin{tabularx}{\textwidth}{lXlX}
\hline
\hline
\textbf{Number} & \textbf{Feature label} &  \textbf{Number} & \textbf{Feature label}   \\
  \\
   \hline
 0 & \textbf{Erythema} & 18 & \textbf{Parakeratosis}   \\ 
 1 & \textbf{Scaling} & 19 & \textbf{Clubbing of the rete ridges} \\ 
 2 & \textbf{Definite borders} & 20 & \textbf{Elongation of the rete ridges}  \\ 
 3 & \textbf{Itching} & 21 & \textbf{Thinning of the suprapapillary epidermis}   \\ 
 4 & \textbf{Koebner phenomenon} & 22 & \textbf{Spongiform pustule}  \\ 
 5 & \textbf{Polygonal papules} & 23 & \textbf{Munro microabcess}   \\ 
 6 & \textbf{Follicular papules} & 24 & \textbf{Focal hypergranulosis}  \\ 
 7 & \textbf{Oral mucosal involvement} & 25 & \textbf{Disappearance of the granular layer}  \\ 
 8 & \textbf{Knee and elbow involvement} & 26 & \textbf{Vacuolisation and damage of basal layer}  \\ 
 9 & \textbf{Scalp involvement} & 27 & \textbf{Spongiosis}   \\ 
 10 & \textbf{Family history} & 28 & \textbf{Saw-tooth appearance of retes}  \\ 
 11 & \textbf{Melanin incontinence} & 29 & \textbf{Follicular horn plug} \\ 
 12 & \textbf{Eosinophils in the infiltrate}  & 30 & \textbf{Perifollicular Parakeratosis}\\
 13 & \textbf{PNL infiltrate}  & 31 & \textbf{Inflammatory monoluclear inflitrate}\\
 14 & \textbf{Fibrosis of the papillary dermis}  & 32 & \textbf{Band-like infiltrate}\\
 15 & \textbf{Exocytosis}  & 33 & \textbf{Age}\\
 16 & \textbf{Ecanthosis} &&\\
 17 & \textbf{Hyperkeratosis}&&\\
 \hline
 \hline
\end{tabularx}
}
\caption{Features in Dermatology dataset. Features considered after data pre-proccessing. The features labeled from 0 to 10, and feature 33 are clinical, whereas from 11 to 32 are histopathological.}
\label{supp_3}
\end{table}

\begin{figure}[!htb]
\centering
\subfloat[RFE.]{\includegraphics[width=0.7\textwidth]{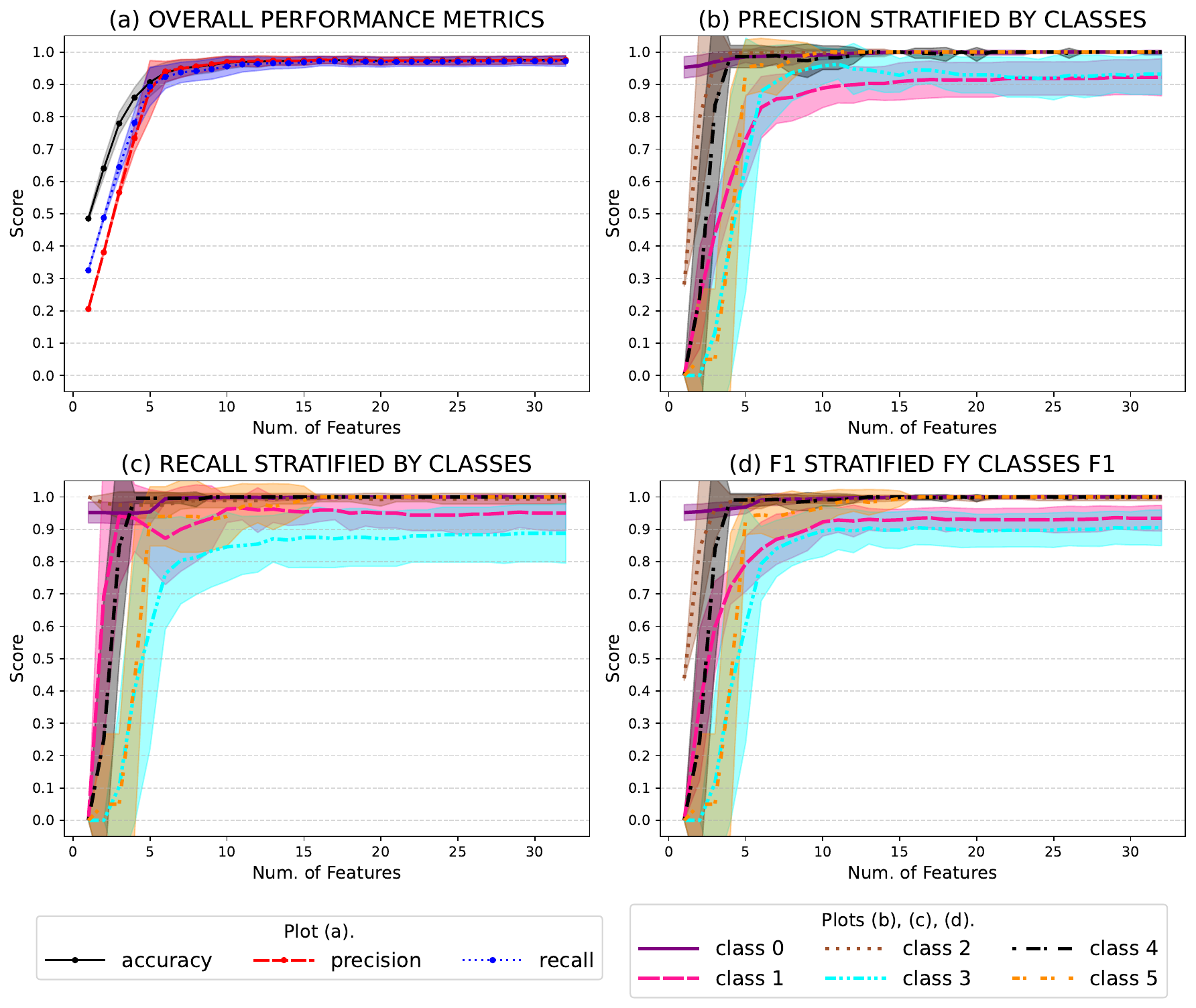}%
\label{fig_2_a}}
\hfil
\subfloat[CRFE.]{\includegraphics[width=0.7\textwidth]{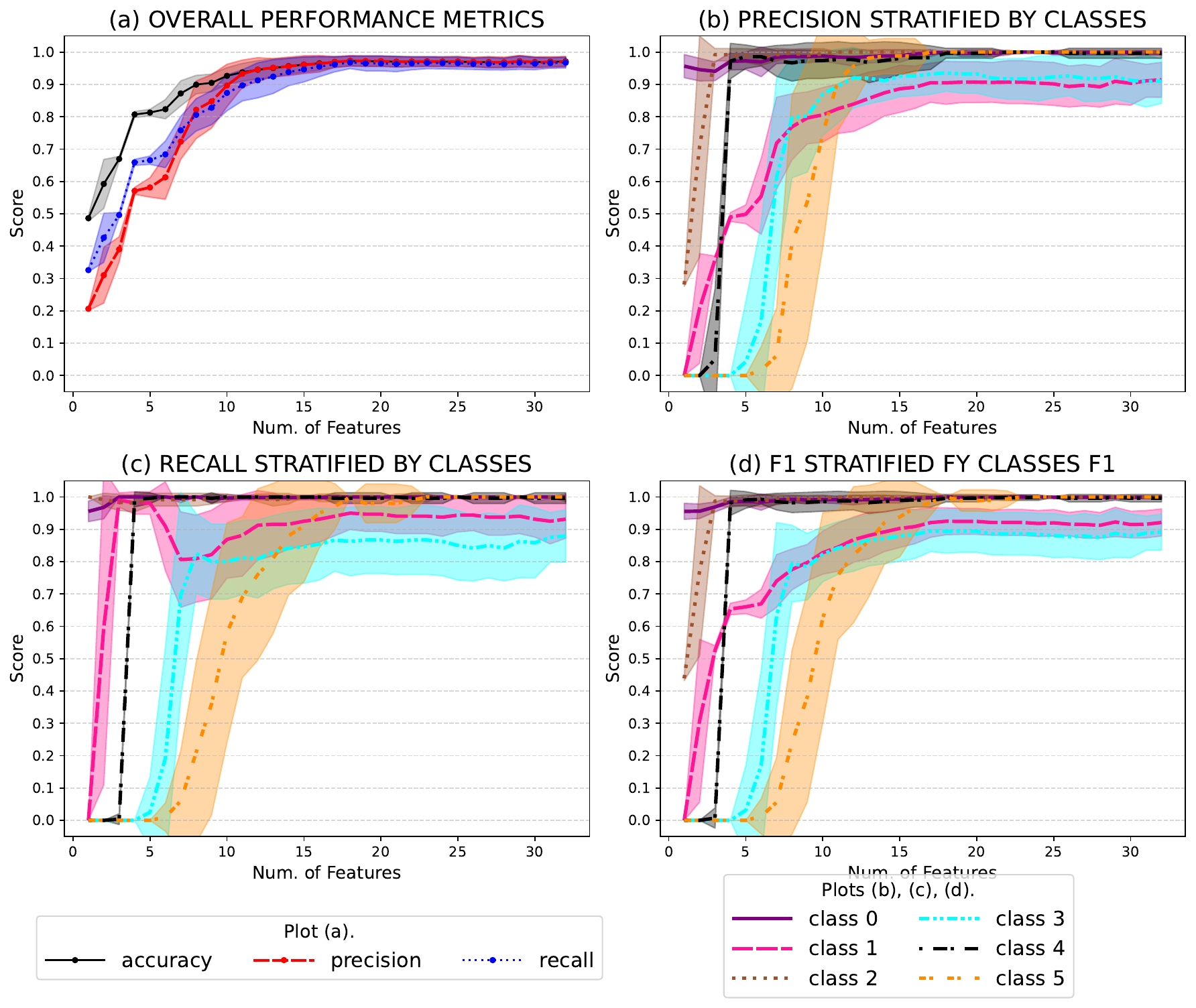}%
\label{fig_2_b}}
\caption{Single-prediction performance metrics for the dermatology dataset. Standard deviation is provided as upper and lower intervals. Plots a-(a) and b-(a) show accuracy, precision, and recall performance metrics achieved by subsets of features selected by RFE and CRFE, respectively. Plots a-(b),(c),(d), and b-(b),(c),(d) show precision, recall, and F1 score performance by class achieved by RFE and CRFE respectively.}
\label{Figure_2}
\end{figure}

\subsection{Myocardial infarction complications}\label{Sec:Myo}
\begin{figure}[!htb]
\centering
\subfloat[RFE.]{\includegraphics[width=0.7\textwidth]{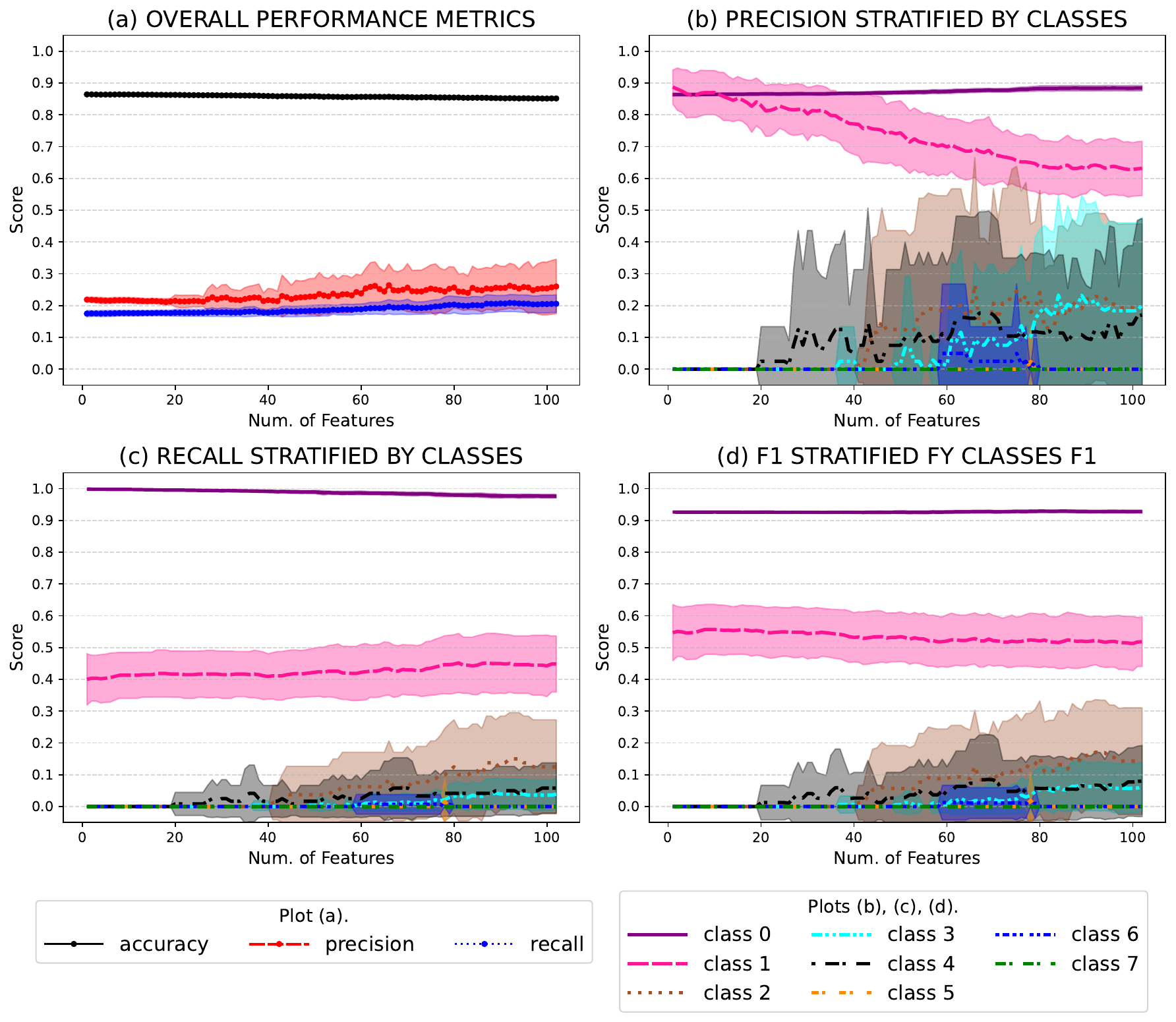}%
\label{fig_3_a}}
\hfil
\subfloat[CRFE.]{\includegraphics[width=0.7\textwidth]{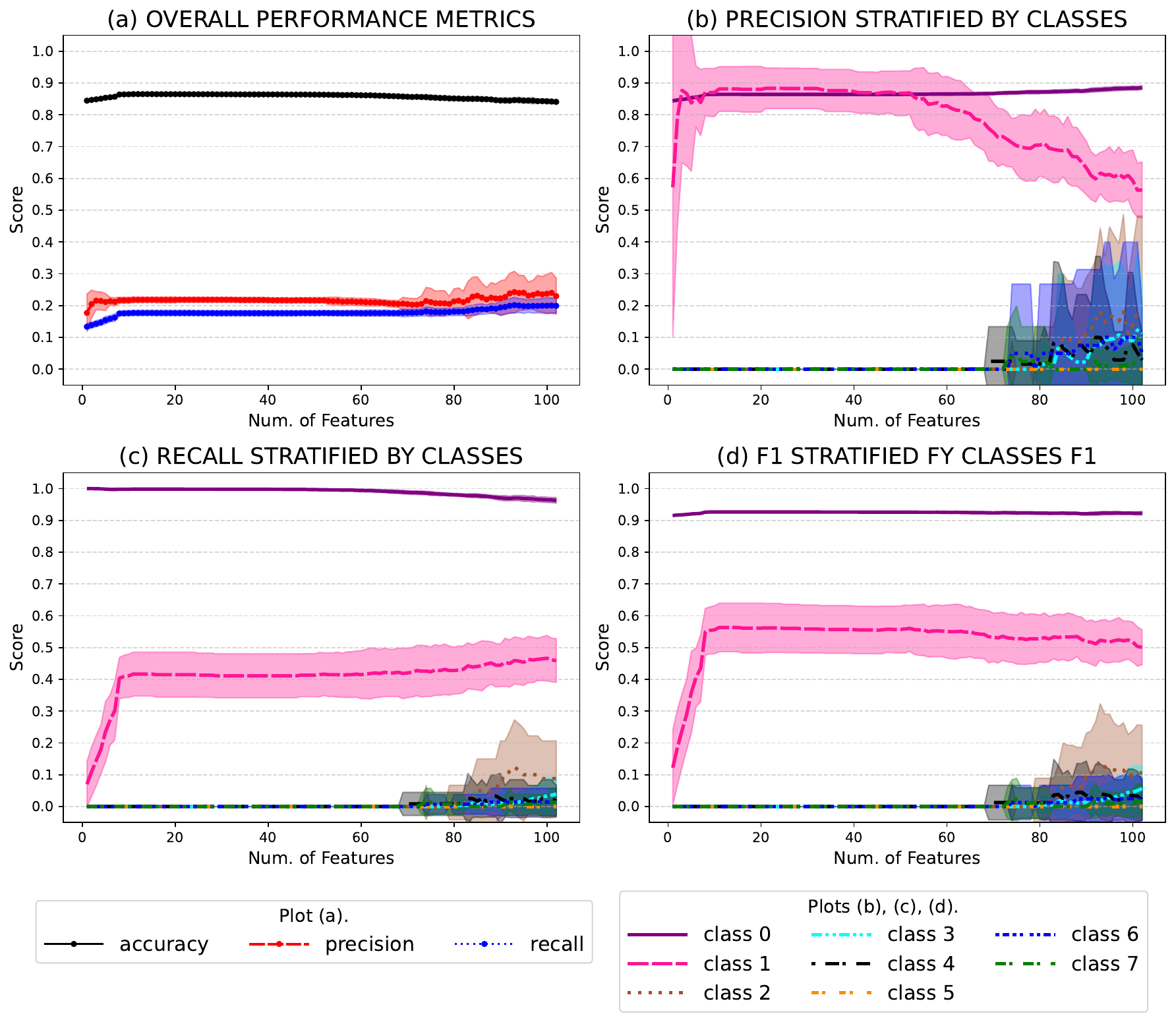}%
\label{fig_3_b}}
\caption{Single-prediction performance metrics for the Myocardial infarction complications dataset. Standard deviation is provided as upper and lower intervals. Plots a-(a) and b-(a) show the accuracy, precision, and recall performance metrics achieved by subsets of features selected by RFE and CRFE respectively. Plots a-(b),(c),(d), and b-(b),(c),(d) show precision, recall, and F1 score performance by class achieved by RFE and CRFE, respectively.}
\label{Figure_3}
\end{figure}

The motivation behind myocardial infarction complications dataset \cite{Golovenkin_2020} were the continuous spread of the disease, with special importance in the urban population of developed countries, as well as the differences between patients in the course of the disease. The feature with patient ID was removed because it did not contain relevant information, other features were also removed due to missing data (features with more than 25\% of missing values), and features numbered from 113 to 124 were removed because were provided as potential targets. The name of the 104 remaining features are shown and numbered in the Table \ref{supp_5}. We refer the reader to the publicly available repository for specific information about the features. Missing data were imputed by a k-nn means algorithm with 5 neighbours. The objective is to predict the feature called Lethal outcome (LET\_IS), distributed as: class \textit{alive} (84.06\%), class \textit{cardiogenic shock} (6.47\%), class \textit{pulmonary edema} (1.06\%), class \textit{myocardial rupture} (3.18\%), class \textit{progress of congestive heart failure} (1.35\%), class \textit{thromboembolism} (0.71\%), class \textit{asystole} (1.59\%) and class \textit{ventricular fibrillation} (1.59\%).
\begin{table}[H]
\centering
\resizebox{1.\textwidth}{!}{
\begin{tabular}{llllll}
\hline
\hline
\textbf{Number} & \textbf{Feature label} &  \textbf{Number} & \textbf{Feature label} &  \textbf{Number} & \textbf{Feature label}   \\
   \hline
 0 & \textbf{AGE} & 35 & \textbf{K\_SH\_POST} & 70 & \textbf{n\_p\_ecg\_p\_12} \\
 1  & \textbf{SEX} & 36 & \textbf{MP\_TP\_POST} & 71 & \textbf{fibr\_ter\_01} \\ 
 2 & \textbf{INF\_ANAM} & 37 & \textbf{SVT\_POST} & 72 & \textbf{fibr\_ter\_02} \\
 3 & \textbf{STENOK\_AN} & 38 & \textbf{GT\_POST} & 73 & \textbf{fibr\_ter\_03} \\
 4 & \textbf{FK\_STENOK} & 39 & \textbf{FIB\_G\_POST)} & 74 & \textbf{fibr\_ter\_05} \\
 5 & \textbf{IBS\_POST} & 40 & \textbf{ant\_im} & 75 &  \textbf{fibr\_ter\_06} \\
 6 & \textbf{GB} & 41 & \textbf{lat\_im} & 76 &  \textbf{fibr\_ter\_07} \\
 7 & \textbf{SIM\_GIPERT} & 42 & \textbf{inf\_im} & 77 & \textbf{fibr\_ter\_08} \\
 8 & \textbf{DLIT\_AG} & 43 & \textbf{post\_im} & 78 & \textbf{GIPO\_K} \\
 9 & \textbf{ZSN\_A} & 44 & \textbf{IM\_PG\_P} & 79 &  \textbf{K\_BLOOD} \\
 10 & \textbf{nr11} & 45 & \textbf{ritm\_ecg\_p\_01} & 80  & \textbf{GIPER\_Na)} \\
 11 & \textbf{nr01} & 46 & \textbf{ritm\_ecg\_p\_02} & 81 &  \textbf{Na\_BLOOD} \\
 12 & \textbf{nr02} & 47 & \textbf{ritm\_ecg\_p\_04} & 82 &  \textbf{ALT\_BLOOD} \\
 13 & \textbf{nr03} & 48 & \textbf{ritm\_ecg\_p\_06} & 83 &  \textbf{AST\_BLOOD} \\
 14 & \textbf{nr04} & 49 & \textbf{ritm\_ecg\_p\_07} & 84 &  \textbf{L\_BLOOD} \\
 15 & \textbf{nr07} & 50 & \textbf{ritm\_ecg\_p\_08} & 85 &  \textbf{ROE} \\
 16 & \textbf{nr08} & 51 & \textbf{n\_r\_ecg\_p\_01} & 86 &  \textbf{TIME\_B\_S} \\
 17 & \textbf{np01} & 52 & \textbf{n\_r\_ecg\_p\_02} & 87 & \textbf{R\_AB\_1\_n} \\
 18 & \textbf{np04} & 53 & \textbf{n\_r\_ecg\_p\_03} & 88 & \textbf{R\_AB\_2\_n} \\
 19 & \textbf{np05} & 54 & \textbf{n\_r\_ecg\_p\_04} & 89 &  \textbf{R\_AB\_3\_n} \\
 20 & \textbf{np07} & 55 & \textbf{n\_r\_ecg\_p\_05} & 90 &   \textbf{NITR\_S} \\
 21 & \textbf{np08} & 56 & \textbf{n\_r\_ecg\_p\_06} & 91 &  \textbf{NA\_R\_1\_n} \\
 22 & \textbf{np09} & 57 & \textbf{n\_r\_ecg\_p\_08} & 92 & \textbf{NA\_R\_2\_n} \\
 23 & \textbf{np10} & 58 & \textbf{n\_r\_ecg\_p\_09} & 93 & \textbf{NA\_R\_3\_n} \\
 24 & \textbf{endocr\_01} & 59 & \textbf{n\_r\_ecg\_p\_10}  & 94 & \textbf{NOT\_NA\_1\_n} \\
 25 & \textbf{endocr\_02} & 60 & \textbf{n\_p\_ecg\_p\_01} & 95 & \textbf{NOT\_NA\_2\_n} \\
 26 & \textbf{endocr\_03} & 61  & \textbf{n\_p\_ecg\_p\_03} & 96 & \textbf{NOT\_NA\_3\_n} \\
 27 & \textbf{zab\_leg\_01} & 62 & \textbf{n\_p\_ecg\_p\_04}  & 97  & \textbf{LID\_S\_n} \\
 28 & \textbf{zab\_leg\_02} & 63 & \textbf{n\_p\_ecg\_p\_05}  & 98 & \textbf{B\_BLOK\_S\_n} \\
 29 & \textbf{zab\_leg\_03} & 64 & \textbf{n\_p\_ecg\_p\_06}  & 99 & \textbf{ANT\_CA\_S\_n} \\
 30 & \textbf{zab\_leg\_04} & 65 & \textbf{n\_p\_ecg\_p\_07}  & 100 & \textbf{GEPAR\_S\_n} \\
 31 & \textbf{zab\_leg\_06} & 66 & \textbf{n\_p\_ecg\_p\_08}  & 101 & \textbf{ASP\_S\_n} \\
 32 & \textbf{S\_AD\_ORIT} & 67 & \textbf{n\_p\_ecg\_p\_09}  & 102 & \textbf{TIKL\_S\_n} \\
 33 & \textbf{D\_AD\_ORIT} & 68 & \textbf{n\_p\_ecg\_p\_10}  & 103 & \textbf{TRENT\_S\_n} \\
 34 & \textbf{O\_L\_POST} & 69 & \textbf{n\_p\_ecg\_p\_11} &  & \\
 \hline
 \hline
\end{tabular}
}
\caption{Myocardial infartion dataset. Features considered after data pre-proccessing.}
\label{supp_5}
\end{table}
Figures \ref{fig_3_a} and \ref{fig_3_b} show that the subsets selected by RFE preserved features relevant for distinguishing between all classes. CRFE quickly attempts to select relevant features for classes \textit{alive} and \textit{cardiogenic shock}. This suggests that a heavy imbalance in classes may affect the performance of the proposed method.\\

Results presented in Figure 4 of the main paper showed that the most relevant features, \textit{i.e.} those present in at least the 80\% of the subset of features were: \textit{sex}, \textit{S\_AD\_ORIT} (the systolic blood pressure), \textit{ritm\_ecg\_p\_01} which represents if the ECG rhythm at the time of admission to hospital is sinus or not, \textit{TIME\_B\_S} which represents the time elapsed from the beginning of the attack of CHD to the hospital, \textit{ANT\_CA\_S\_n} stands for the use of calcium channel blockers in the ICU, and finally \textit{ASP\_S\_n} if acetylsalicylic acid was used in the ICU. Accordingly with supplementary Figures \ref{fig_3_a} and \ref{fig_3_b}, these features were postulated to be the most relevant to predict a fatal outcome or not.

\subsection{Similarity study}\label{Sec:Sim}
We show the plots supporting the similarity study between the subset of features selected by RFE and CRFE.
Three consistency indexes were used for the study. Firstly, we need to define the standard Jaccard index as 
\begin{equation}
    I_J = \frac{|A \cap B|}{|A \cup B|},
\end{equation}
where \(A\) and \(B\) are two sets such as \(|A| = |B|\). Through this index, we compared the two subset of features selected by both methods that (i) were selected using the same random seed and (ii) had the same size. The average and the standard deviation are provided in Supplementary Figure \ref{stab_2}.
The novel index defined in Equation (15) of the main document was suitable to compare subsets of features produced by different feature selection methods having the same cardinality. The new index did not need to be averaged through the multiple runs because it takes into account all the subsets generated at the same time. This was possible because of the \(n/2 +1\) condition, which is the minimum number of common features required to ensure that a feature is present in at least two subsets generated from two different feature selection methods. See Supplementary Figure \ref{stab_2}.
The last index to study similarity is the Kuncheva index \cite{Kuncheva_2007}. This index is also based on the cardinality of the intersection between sets of elements, but introduces a correction for agreements by chance. It was defined as:
\begin{equation}
    \label{kuncheva}
    I_K = \frac{rs-\kappa^2}{\kappa (s-\kappa)},
\end{equation}
where \(r = |A \cap B|\), \(|A| = |B| = \kappa\) and \(0\leq \kappa \leq|\mathcal{X}| = s\).
\begin{figure*}[!htb]
\centering
\subfloat[Jaccard and weighted Jaccard indexes.]{\includegraphics[width=0.7\textwidth]{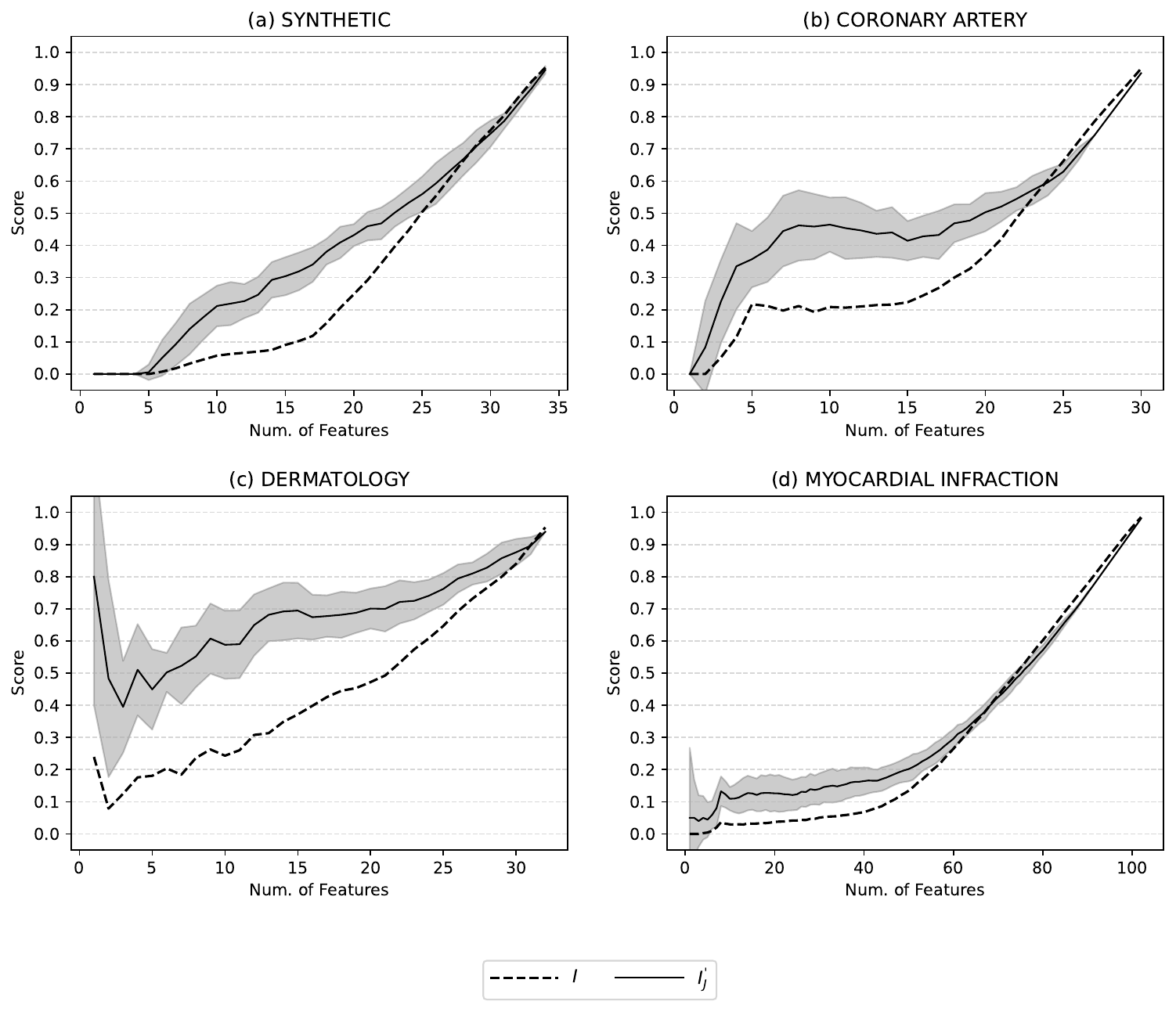}%
\label{stab_2}}
\hfil
\subfloat[Kuncheva index.]{\includegraphics[width=0.7\textwidth]{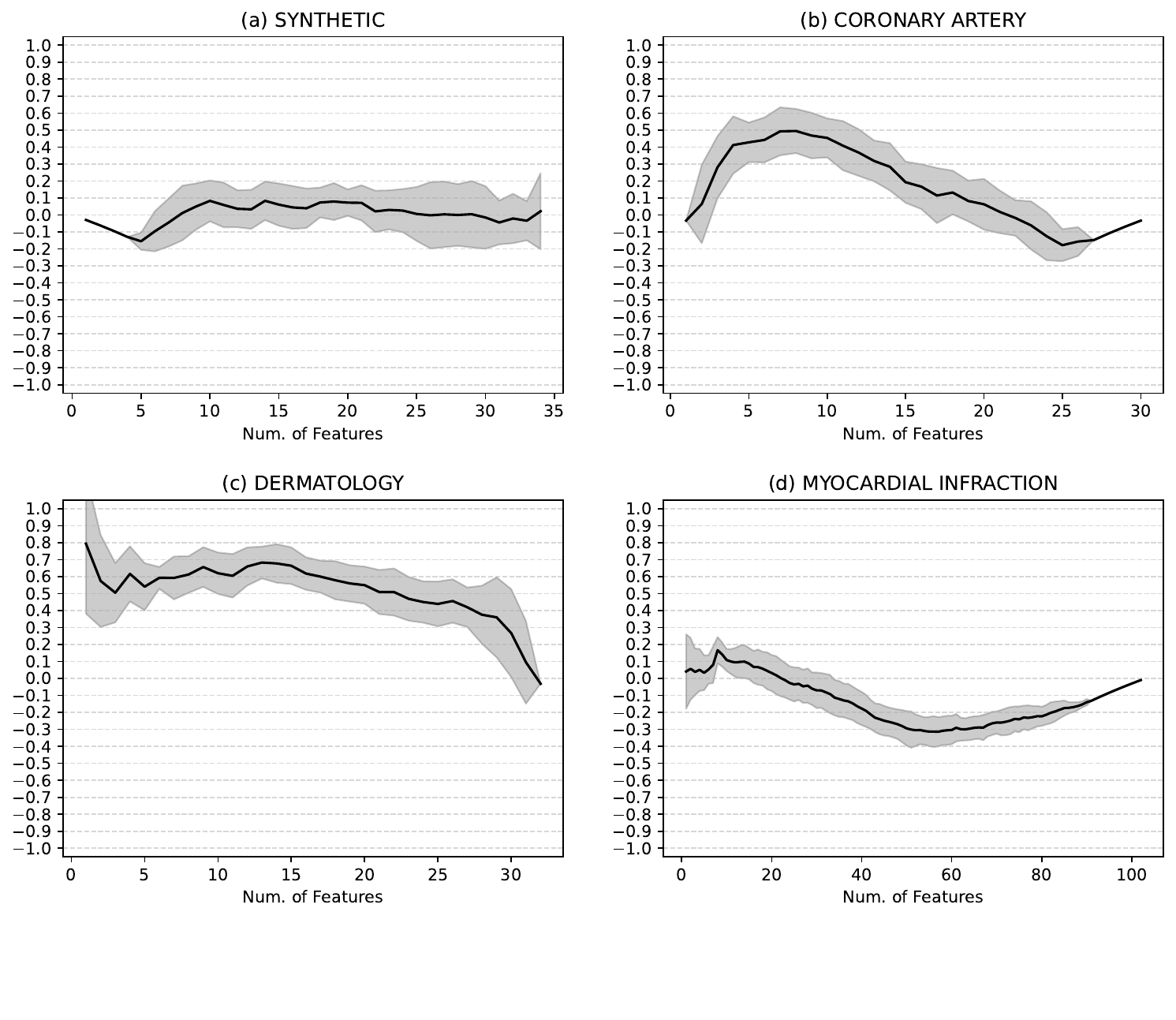}%
\label{KUN}}
\caption{Study of the common features selected by both feature selection methods for each dataset. The subsets of features compared using the Jaccard and new index in Plot a and Kuncheva index in Plot b were those generated with the same random numbers and size, but by different feature selection method. }
\label{consistency_analysis_2}
\end{figure*}
The index was used in the same way as Jaccard in Figure \ref{stab_2}. The averaged results and standard deviations are shown in Figure \ref{KUN}.

\subsection{Stopping criteria}\label{Sec:Stop}

\begin{figure}[htb]
    \centering
    \includegraphics[width=1\textwidth]{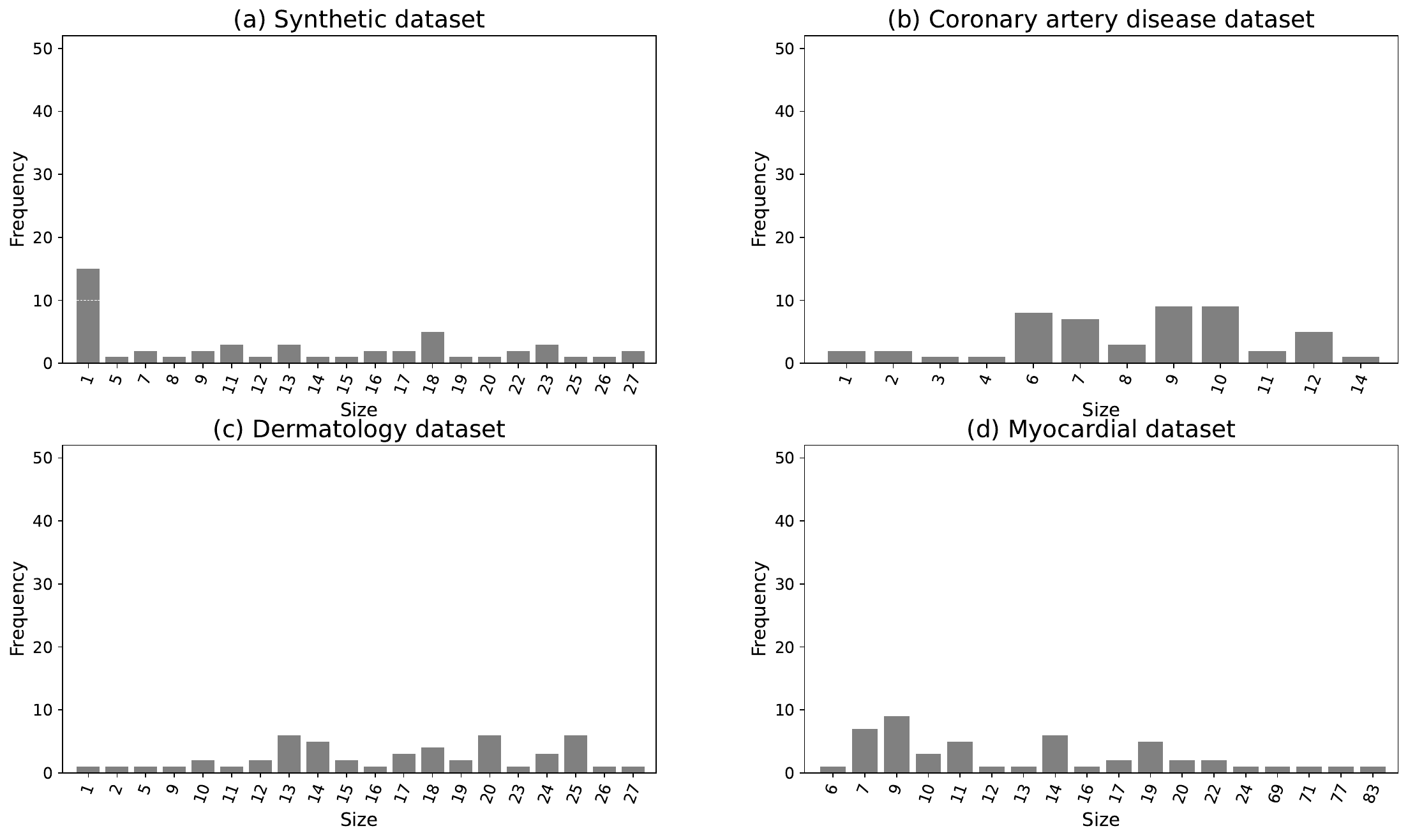}
    \caption{Distribution of sizes. Subsets of features selected by the \(\beta\)-based stopping criteria in Subsection IV-E of the main document.}
    \label{frec_sizes}
\end{figure}

Figure \ref{frec_sizes} exposes the distribution of subset sizes when using the \(\beta\)-based stopping criterion in 50 independent runs. Figure \ref{frec_sizes}-(a) shows how the most frequent subsets were made of a single feature. Figure \ref{frec_sizes}-(b) shows how the subset sides produced by the coronary artery disease dataset range from 8 to 10 features, corresponding with the desirable size based on the analysis conducted in the main document. In Figure \ref{frec_sizes}-(c), more counts were measured for sizes between 13 and 14 features, as well as between 17 to 20. According to the main document, sizes ranged from 17 to 25 yield superior results. Lastly, Figure \ref{frec_sizes}-(d) exhibits more presence of sizes of 7 and 9 features. 

\bibliographystyle{plain}
\bibliography{Bibliografia.bib}

\end{document}